\documentclass[mathematics,article,accept,moreauthors,pdftex]{Definitions/mdpi} 
\makeatletter
\def\UrlAlphabet{%
      \do\a\do\b\do\c\do\d\do\e\do\f\do\g\do\h\do\i\do\j%
      \do\k\do\l\do\m\do\n\do\o\do\p\do\q\do\r\do\s\do\t%
      \do\u\do\v\do\w\do\x\do\y\do\z\do\A\do\B\do\C\do\D%
      \do\E\do\F\do\G\do\H\do\I\do\J\do\K\do\L\do\M\do\N%
      \do\O\do\P\do\Q\do\R\do\S\do\T\do\U\do\V\do\W\do\X%
      \do\Y\do\Z}
\def\UrlDigits{\do\1\do\2\do\3\do\4\do\5\do\6\do\7\do\8\do\9\do\0}
\g@addto@macro{\UrlBreaks}{\UrlOrds}
\g@addto@macro{\UrlBreaks}{\UrlAlphabet}
\g@addto@macro{\UrlBreaks}{\UrlDigits}
\makeatother

\usepackage[utf8]{inputenc}
\usepackage[english]{babel}

\usepackage{amsmath}
\usepackage{graphicx}
\usepackage{caption}
\usepackage[labelformat=simple]{subcaption}

\usepackage{enumitem}
\usepackage{pdflscape}

\usepackage[finalnew]{trackchanges}

\firstpage{1} 
\pubvolume{1}
\issuenum{1}
\articlenumber{0}
\pubyear{2022}
\copyrightyear{2022}
\externaleditor{{Academic Editor(s): Zhao Kang}} 

\datereceived{9 November 2022} 
\dateaccepted{30 November 2022} 
\datepublished{} 
\hreflink{https://doi.org/} 

\Title{A Comparison of Several AI Techniques for Authorship Attribution on Romanian Texts}

\TitleCitation{A Comparison of Several AI Techniques for Authorship Attribution on Romanian Texts}

\Author{Sanda-Maria Avram 
$^{1,}$*\orcidA{} and Mihai Oltean $^{2}$\orcidB{}}

\AuthorNames{Sanda-Maria Avram and Mihai Oltean}

\AuthorCitation{Avram, S.-M.; Oltean, M.}

\address[1]{$^{1}$ \quad Faculty of Mathematics and Computer Science, Babeș-Bolyai  University, 400084 Cluj-Napoca, Romania 
\\
$^{2}$ \quad Independent Researcher, Cugir, 515600, Romania; 
mihai.oltean@gmail.com}

\corres{Correspondence: sanda.avram@ubbcluj.ro}

 \abstract{Determining the author of a text is a difficult task. Here, we compare multiple Artificial Intelligence techniques for classifying literary texts written by multiple authors by taking into account a limited number of speech parts (prepositions, adverbs, and conjunctions). We also introduce a new dataset composed of texts written in the Romanian language on which we have run the algorithms. The compared methods are artificial neural networks, multi-expression programming, k-nearest neighbour, support vector machines, and decision trees with C5.0. Numerical experiments show, first of all, that the problem is difficult, but some algorithms are able to generate 
 acceptable error rates on the test set.}


\begin{document}



\section{Introduction}
Automated authorship attribution (AA) is defined in~\cite{oliveira2013comparing} as the task of determining authorship of an unknown text based on the textual characteristics of the text itself. Today the AA is useful in a plethora of fields: from the educational and research domain to detect plagiarism~\cite{stamatatos2011plagiarism} to the justice domain to analyze evidence on forensic cases~\cite{koppel2008authorship} and cyberbullying~\cite{xu2012fast}, to the social media~\cite{sinnott2021linking,zhang2016authorship} to detect compromised accounts~\cite{barbon2017authorship}. 

Most approaches in the area of artificial intelligence treat the AA problem by using simple classifiers (e.g., linear SVM or decision tree) \change[E.E.]{having}{that have} bag-of-words (character n-grams) as features or other conventional feature sets~\cite{kestemont2021overview, kestemont2018overview}. Although deep neural learning was already used for natural language processing (NLP), the adoption of such strategies for authorship identification occurred later. In recent years, pre-trained language models (such as BERT and GPT-2) have been used for finetuning and accuracy improvements~\cite{kestemont2021overview, tyo2022state, barlas2021transfer}. 

The challenges in solving the AA problem can be grouped into three main groups ~\cite{kestemont2021overview}:
\begin{enumerate}
  \item The lack of large-scale datasets;
  \item The lack of methodological diversity;
  \item The ad hoc nature of authorship.
\end{enumerate}

The availability of large-scale datasets has improved in recent years as large datasets have become widespread~\cite{PANdatasets, tatmanBlog}. 
Other issues that relate to the datasets are the language in which the texts are written, the domain, the topic, and the writing environment. Each of these aspects has its own particularities. From the language perspective, the issue is that most available datasets consist of texts written in English. There is PAN18~\cite{kestemont2018overview} for \remove[E.E.]{five languages:}English, French, Italian, Polish, and Spanish; or PAN19~\cite{kestemont_mike_2019_3530313} for \remove[E.E.]{four languages:}English, French, Italian, and Spanish. However, there are not very many datasets for other languages and this is \change[E.E.]{very important}{crucial} as there are particularities that pertain to the language~\cite{al2020ensemble}.

The methodological diversity has also improved in recent years, as it is detailed in~\cite{kestemont2021overview}. However, the ad hoc nature of authorship is a more difficult issue, as a set of features that differentiates one author from the rest may not work for another author due to the individuality aspect of different writing styles. Even for one author, the writing style can evolve or change over a period of time, or it can differ depending on the context (e.g., the domain, the topic, or the writing environment). Thus, modeling the authorial writing style has to be carefully considered and needs to be tailored to a specific set of authors~\cite{kestemont2021overview}. Therefore, selecting a distinguishing set of features is a \remove[E.E.]{very}challenging task.

We propose a new dataset named ROST (ROmanian Stories and other Texts) 
as there are few available datasets that contain texts written in Romanian~\cite{romDatasets}. The existing datasets are small, on obscure domains, or translated from other languages. Our dataset consists of 400 texts written by 10 authors. We have elements that pertain to the intended heterogeneity of the dataset such as: 
\begin{itemize}
  \item Different text types: stories, short stories, fairy tales, novels, articles, and sketches;
  \item Different number of texts per author: ranging from 27 to 60;
  \item Different sources: texts are collected from 4 different websites\remove[A.]{ (sometimes there are different websites as sources for the same author)}; 
  \item Different text lengths: ranging from 91 to 39,195 words per text;
  \item Different periods: the time period in which the considered texts were written spans over 3 centuries, which introduces diachronic developments;
  \item Different mediums: \change[A.]{from texts printed on paper and }{texts were} written with the intention of being read from paper medium (most of the considered authors) to online (two contemporary authors). This aspect considerably changes the writing style, as shorter sentences and shorter words are used online, and \change[E.E.]{also}{they also contain} more adjectives and pronouns~\cite{wang2021mode}.
\end{itemize}

As our set is heterogeneous (as described above) from multiple perspectives, the \change[A.]{problem of author attribution}{authorship attribution} is even more difficult. We investigate this classification problem by using five different techniques from the artificial intelligence area:
\begin{enumerate}
  \item Artificial neural networks (ANN);
  \item Multi-expression programming (MEP);
  \item K-nearest neighbor (k-NN);
  \item Support vector machine (SVM);
  \item Decision trees (DT) with C5.0.
\end{enumerate}

For each of these methods, we investigate different scenarios by varying the number and the type of some features \remove[E.E.]{in order} to determine the context in which they obtain the best results. The aim of our investigations is twofold. On one side, the result of this investigation is to determine which method performs best while working on the same data. On another side, we try to find out the proper number and type of features that best classify the authors on this specific dataset. 

\remove[A.]{The text classification process that we followed in our work is depicted in Figure \ref{fig:rez-discuss}. Another form of such a process is also presented in~\cite{halteren05new}.}

The paper is organized as follows:
\begin{itemize}
  \item Section~\ref{related_work} 
  describes the AA state of the art by using methods from artificial intelligence; details the entire prerequisite process to be considered before applying the specific AI algorithms (highlighting possible ``stylometric features'' to be considered); provides a table with some available datasets; presents a number of AA methods already proposed; describes the steps of the attribution process; 
  presents an overview and a comparison of AA state-of-the-art 
  methods.
  \item Section~\ref{proposed_dataset} details the specific particularities (e.g., in terms of size, sources, time frames, types of writing, and writing environments) of the database we are proposing and we are going to use, and the building and scaling/pruning process of the feature set.
  \item Section~\ref{compared_methods} introduces the five methods we are going to use in our investigation.
  \item Section~\ref{numerical_experiments} \remove[A.]{provides the parameters used by each method and some preliminary investigations to determine the best values of some of these parameters;}presents the results and interprets them, making a comparison between the five methods and the different sets of features used; 
  measures the results by using metrics that allow a comparison with the results of other state-of-the-art methods.
  \item Section~\ref{future_work} concludes with final remarks on the work and provides future possible directions and investigations.
\end{itemize}

\section{Related Work}
\label{related_work}

The AA detection can be modeled as a classification problem. The starting premise is that each author has a stylistic and linguistic ``fingerprint'' in their work~\cite{halteren05new}. Therefore, in the realm of AI, this means extracting a set of characteristics, which can be identified in a large-enough writing sample \cite{kestemont2021overview}. 

\subsection{Features}
\label{related_work-features}
\change[E.E.]{The characteristics by which an author's style can be identified are called ``stylometric features'' that can be quantified and learned \cite{grondahl2019text}. They can be }{\emph{Stylometric features} are the characteristics that define an author's style. They can be quantified, learned}~\cite{grondahl2019text}, and classified into five groups~\cite{stamatatos2009survey}:
\begin{enumerate}
  \item Lexical 
    (the text is viewed as a sequence of tokens grouped into sentences, with each token corresponding to a word, number, or punctuation mark):
    \begin{itemize}
      \item Token-based 
      (e.g., word length, sentence length, etc.);
      \item Vocabulary richness (i.e., attempts to quantify the vocabulary diversity of a text);
      \item Word frequencies (e.g., the traditional ``bag-of-words'' representation~\cite{sebastiani2002machine} in which texts become vectors of word frequencies disregarding contextual information, i.e., the word order);
      \item Word n-grams (i.e., sequences of n contiguous words also known as \emph{word collocations});
      \item Errors (i.e., intended to capture the idiosyncrasies of an author’s style) (requires orthographic spell checker).
    \end{itemize} 
  \item Character (the text is viewed as a sequence of characters):
    \begin{itemize}
      \item Character types (e.g., letters, digits, etc.) (requires character dictionary);
      \item Character n-grams 
      (i.e., considers all sequences of $n$ consecutive characters in the texts; $n$ can have a variable or fixed length);
      \item Compression methods (i.e., the use of a compression model acquired from one text to compress another text; compression models are usually based on repetitions of character sequences).
    \end{itemize} 
  \item Syntactic (text-representation which considers syntactic information):
    \begin{itemize}
      \item Part-of-speech (POS) (requires POS tagger---a tool that assigns a tag of morpho-syntactic information to each word-token based on contextual information);
      \item Chunks (i.e., phrases);
      \item Sentence and phrase structure (i.e., a parse tree of each sentence is produced);
      \item Rewrite rules frequencies (these rules express part of the syntactic analysis, helping to determine the syntactic class of each word as the same word can have different syntactic values based on different contexts);
      \item Errors (e.g., sentence fragments, run-on sentences, mismatched use of tenses) (requires syntactic spell checker).
    \end{itemize} 
  \item Semantic (text-representation which considers semantic information):
    \begin{itemize}
      \item Synonyms (requires thesaurus);
      \item Semantic dependencies.
    \end{itemize} 
  \item Application-specific (the text is viewed from an application-specific perspective to better represent the nuances of style in a given domain):
    \begin{itemize}
      \item Functional (requires 
      specialized dictionaries);
      \item Structural (e.g., the use of greetings and farewells in messages, types of signatures, use of indentation, and paragraph length); 
      \item Content-specific (e.g., content-specific keywords);
      \item Language-specific.
    \end{itemize} 
\end{enumerate}

The lexical and character \change[E.E.]{-specific characteristics }{features} are simpler because they view the text as a sequence of word-tokens or characters, not requiring any linguistic analysis, in contrast with the syntactic and semantic characteristics, which do. The application-specific characteristics are restricted to certain text domains or languages. 

A simple and successful feature selection, based on lexical characteristics, is made by using the top of the $N$ most frequent words from a corpus containing texts of the candidate author. Determining the best value of $N$ was the focus of numerous studies, starting from 100~\cite{burrows1987word}, and reaching 1000~\cite{stamatatos2006authorship}, or even all words that appear at least twice in the corpus~\cite{madigan2005author}. It was observed, that depending on the value of $N$, different types of words (in terms of content specificity) \change[E.E.]{are}{make up} the majority. Therefore, when the size of $N$ falls within dozens, the most frequent words of a corpus are closed-class words (i.e., articles, prepositions, etc.), while when $N$ exceeds a few hundred words, open-class words (i.e., nouns, adjectives, verbs) are the majority~\cite{stamatatos2009survey}.

Even though the word n-grams approach comes as a solution to keeping the contextual information, i.e., the word order, which is lost in the word frequencies (or ``bag-of-words'') approach, the classification accuracy is not always better~\cite{coyotl2006authorship,sanderson2006short}.

The main advantage of \change[E.E.]{feature selection based on character-specific characteristics }{character feature selection} is that they pertain to any natural language and corpus. Furthermore, even the simplest in this category (i.e., character types) proved to be useful to quantify the writing style~\cite{grieve2007quantitative}.

The character n-grams have 
the advantages of capturing the nuances of style and being tolerant to noise (e.g., grammatical errors or making strange use of punctuation), and the disadvantage is that they capture redundant information~\cite{stamatatos2009survey}.

The \change[E.E.]{feature selection based on syntactic-specific characteristics}{syntactic feature selection} requires the use of Natural Language Processing (NLP) tools \remove[E.E.]{in order} to perform a syntactic analysis of texts, and they are language-dependent. Additionally, being a method that requires complex text processing, noisy datasets may be produced due to unavoidable errors made by the parser~\cite{stamatatos2009survey}.

For \change[E.E.]{feature selection based on semantic-specific characteristics}{semantic feature selection} an even more detailed text analysis is required for extracting stylometric features. Thus, the measures produced may be less accurate as more noise may be introduced while processing the text. NLP tools are used here for sentence splitting, POS tagging, text chunking, and partial parsing. However, complex tasks, such as full syntactic parsing, semantic analysis, and pragmatic analysis, are hard to be achieved for an unrestricted text~\cite{stamatatos2009survey}. 

A comprehensive survey of the state of the art in stylometry is conducted in~\cite{neal2017surveying}.

The most common approach used in AA 
is to extract features that have a high discriminatory potential~\cite{zhang2014authorship}. There are multiple aspects that have to be considered in AA for selecting the appropriate set of features. Some of them are the language, the literary style (e.g., poetry, prose), the topic addressed by the text (e.g., sports, politics, storytelling), the length of the text (e.g., novels, tweets), the number of text samples, and the number of considered features. For instance, \change[E.E.]{some feature types such as those based on lexical and character characteristics}{lexical and character features}, although more simple, can considerably increase the dimensionality of the feature set~\cite{stamatatos2009survey}. Therefore, feature selection algorithms can be applied to reduce the dimensionality of such feature sets~\cite{forman2003extensive}. This also helps the classification algorithm to avoid overfitting on the training data.

Another prerequisite for the training phase \change[E.E.]{consists in}{is} deciding whether the training \change[E.E.]{text is treated}{texts are processed} individually or cumulatively (per author). From this perspective, the following two \remove[E.E.]{classes of}approaches can be distinguished~\cite{stamatatos2009survey}:
\begin{enumerate}  
    \item \emph{Instance-based approach} (i.e., each training text is individually represented as a separate instance in the training process \remove[E.E.]{in order}to create a reliable attribution model);
    \item \emph{Profile-based approach} (i.e., a cumulative representation of an author’s style, also known as the author’s profile, is extracted by concatenating all available training texts of one author into one large file \remove[E.E.]{in order}to \change[E.E.]{disregard}{flatten} differences between texts).
\end{enumerate}

Efstathios Stamatatos 
offers in~\cite{stamatatos2009survey} a comparison between the two aforementioned approaches. 

\newpage
\subsection{Datasets}
\label{related_work-datasets}

\add[A.]{In Table}~\ref{tab:datasets}, \add[A.]{we present a list of datasets used in AA investigations.}

\add[A.]{There is a large variation between the datasets. In terms of language, there are usually datasets with texts that are written in one language, and there are a few that have texts written in multiple languages. However, most of the available datasets contain texts written in English.}

\add[A.]{The \emph{Size} column is generally the number of texts and authors that have been used in AA investigations. For example, PAN11 and PAN12 have thousands of texts and hundreds of authors. However, in the referenced paper, only a few were used. The datasets vary in the number of texts from hundreds to hundreds of thousands, and in terms of the number of authors, from tens to tens of thousands.}

\begin{table}[H]
    \caption{Datasets used for author attribution detection; \emph{Author(s)} are names of individuals who created the dataset (for a group consisting of more than two, only the name of the first person is provided in the list, followed by ``et~al.''); \emph{Paper} is the first paper that introduced that dataset or that is recommended by its creator(s) to be used for citing the dataset; \emph{Language} is the language in which the texts in the database were written; \emph{Size} is the dimension of the dataset; \emph{Features} stands for the types of features that can be or were used on that specific dataset (however, the information here is only indicative and should not be taken literally); \emph{No. of features}, is also more an indicative value for possible feature set dimensions; \emph{Name or link} provides the name by which that specific dataset is known and, when available, links are provided.\label{tab:datasets}}
    \begin{adjustwidth}{-\extralength}{0cm}
    \begin{tabularx}{\fulllength}{CCCCCCCC}
        \toprule
  	\textbf{Author(s)}&\textbf{Paper}&\textbf{Language}&\textbf{Size}&\textbf{Features}&\textbf{No. of Features}& \textbf{Name or Link}\\
  	\midrule 
   
        Sanda-Maria Avram 
        & this paper & Romanian & {400 texts; 10~authors} & conjunctions, prepositions, and adverbs & $27 + 85 + 670 = 782$ & \href{https://www.kaggle.com/datasets/sandamariaavram/rost-romanian-stories-and-other-texts}{ROST}\\
        \midrule

        Shlomo Argamon and Patrick Juola & \cite{argamon2011clef} & English & {42 literary texts and novels; 14~authors} & words, characters, n-grams & $> 3000$ & \href{https://pan.webis.de/data.html\#pan12-attribution}{PAN11} \cite{argamon_shlomo_2011_3713246}\\
        \midrule

        Patrick Juola & \cite{juola2012} & English & {42 literary texts and novels; 14~authors} & words, characters, n-grams & $> 3000$ & \href{https://pan.webis.de/data.html\#pan12-attribution}{PAN12}\\
        \midrule

        Mike Kestemont et~al. & \cite{ kestemont2018overview} & English, French, Italian, Polish, Spanish. & 2000 fanfiction texts; 20~authors & char n-gram, word n-gram, stylistic, tokens & $> 500$ & \href{https://pan.webis.de/data.html\#pan18-authorship-attribution}{PAN18} \cite{PAN18}\\
        \midrule

        Mike Kestemont et~al. & \cite{kestemont2019clef} & English, French, Italian, Spanish. & 2997 cross-topic fanfiction texts; 36~authors & char n-gram, word n-gram, tokens & $> 300$ & \href{https://pan.webis.de/data.html\#pan19-authorship-attribution}{PAN19} \cite{kestemont_mike_2019_3530313}\\
        \midrule

        Mike Kestemont et~al. & \cite{kestemont2021overview} & English & {443 000 
        cross-topic fanfiction texts;   278 000
        ~authors} & char n-gram, word n-gram, tokens & $> 300$ & \href{https://pan.webis.de/data.html\#pan20-authorship-verification}{PAN20} \cite{kestemont2020clef}\\
        \midrule

        Daniel Pavelec et~al. & \cite{pavelec2008using} & Portuguese & 600 articles; 20~authors & conjunctions and adverbs & $77 + 94 = 171$ & $-$\\
        \midrule

        Paulo Varela et~al. & \cite{varela2010verbs} & Portuguese & 600 articles; 20~authors & conjunctions, adverbs, verbs and pronouns & 77 + 94 + 50 + 91 = 312 & $-$\\
        \midrule

        Yanir Seroussi et~al. & \cite{seroussi2011ghosts} & English & {1342 legal documents; 3~authors} & unigrams, n-grams & $> 2000$ & Judgment\\
        \midrule

        Yanir Seroussi et~al. & \cite{seroussi2010collaborative} & English & {79 550 
        movie reviews; 62~authors} & unigrams, n-grams & $> 200$ & IMDb62\\
 
        \bottomrule
    \end{tabularx} 
    \end{adjustwidth} 
\end{table}

\begin{table}[H]\ContinuedFloat
    \caption{{\em Cont.}}
    \begin{adjustwidth}{-\extralength}{0cm}
    \begin{tabularx}{\fulllength}{CCCCCCCC}
        \toprule
	\textbf{Author(s)}&\textbf{Paper}&\textbf{Language}&\textbf{Size}&\textbf{Features}&\textbf{No. of Features}& \textbf{Name or Link}\\
        \midrule

        Yanir Seroussi et~al. & \cite{seroussi2011personalised} & English & 204 809 
        posts, {66 816 
        reviews; 22 116 
        users} & unigrams, n-grams & $> 1000$ & IMDB1M\\
        \midrule

        Efstathios Stamatatos & \cite{stamatatos2008author} & English & {5000 newswire documents; 50~authors} & unigrams, n-grams & $> 500$ & CCAT50\\
        \midrule

        Efstathios Stamatatos & \cite{stamatatos2012robustness} & English & {444 articles, book reviews; 13~authors} & words, characters, 3-grams & $> 10 000$ 
        & Guardian10\\
        \midrule

        Efstathios Stamatatos & & English & {1000 CCTA industry news; 10~authors} & words, characters, 3-grams & $> 500$ & \href{https://pan.webis.de/data.html\#c10-attribution}{C10}\\
        \midrule

        Efstathios Stamatatos & & English & {5000 CCTA industry news; 50~authors} & words, characters, n-grams & $> 500$ & \href{https://pan.webis.de/data.html\#c50-attribution}{C50}\\
        \midrule

        Jonathan Schler et~al. & \cite{schler2006effects} & English & {over 600 000 
        posts; 19 000 
        bloggers} & tokens, n-grams & $> 200$ & \href{https://www.kaggle.com/datasets/rtatman/blog-authorship-corpus}{Blogs50}\\
        \midrule

        Jade Goldstein et~al. & \cite{goldstein2008creating} & English & {756 documents; 21~authors} & tokens, n-grams & $> 600$ & CMCC\\
        \midrule

        Project Gutenberg & & English & {29 000 
        books; 4500~authors} & tokens, n-grams & $> 60 000$ 
        & \href{https://www.gutenberg.org/}{Gutenberg}\\
        \bottomrule
    \end{tabularx}
    \end{adjustwidth}
\end{table}

\subsection{Strategies}
\label{related_work-strategies}

According to~\cite{mironczuk2018recent}, the entire process of text classification occurs in 6 stages:
\begin{enumerate}
    \item Data acquisition (from one or multiple sources);
    \item Data analysis and labeling;
    \item Feature construction and weighting;
    \item Feature selection and projection;
    \item Training of a classification model;
    \item Solution evaluation.
\end{enumerate}

The classification process initiates with data acquisition, which \change[E.E.]{results in obtaining}{is used to create} the dataset. There are two strategies for the analysis and labeling of the dataset~\cite{mironczuk2018recent}: labeling groups of texts (also called \emph{multi-instance learning})~\cite{liu2018selective}, or assigning a label or labels to each text part (by using supervised methods)~\cite{cunningham2008supervised}. To yield the appropriate data representation required by the selected learning method, first, the features are selected and weighted~\cite{mironczuk2018recent} according to the obtained labeled dataset. 
Then, the number of features is reduced \change[E.E.]{(a phase which can be seen as a data compression; by selecting only the most important features)}{by selecting only the most important features} and projected onto a lower dimensionality. There are two different representations of textual data: \emph{vector space representation}~\cite{manning2008introduction} where the document is represented as a vector of feature weights, and \emph{graph representation}~\cite{mihalcea2011graph} where the document is modeled as a graph (e.g., nodes represent words, whereas edges represent the relationships between the words). In the next stage, different learning approaches are used to train a classification model. Training algorithms can be grouped into different approaches~\cite{mironczuk2018recent}: \emph{supervised}~\cite{cunningham2008supervised} (i.e., any machine learning process), \emph{semi-supervised}~\cite{altinel2017instance} (also known as self-training, co-training, learning from the labeled and unlabeled data, or transductive learning), \emph{ensemble}~\cite{lochter2016short} (i.e., training multiple classifiers and considering them as a “committee” of decision-makers), \emph{active}~\cite{hu2016active} (i.e., the training algorithm has some role in determining the data it will use for training), \emph{transfer}~\cite{weiss2016survey} (i.e., the ability of a learning mechanism to improve the performance for a current task after having learned a different but related concept or skill from a previous task; also known as \emph{inductive transfer} or \emph{transfer of knowledge across domains}), or \emph{multi-view learning}~\cite{zhao2017multi} (also known as \emph{data fusion} or \emph{data integration} from multiple feature sets, multiple feature spaces, or diversified feature spaces that may have different distributions of features).

By providing probabilities or weights, the trained classifier is then able to decide a class for each input vector. Finally, the classification process is evaluated. The performance of the classifier can be measured based on different indicators~\cite{mironczuk2018recent}: precision, recall, accuracy, F-score, specificity, area under the curve (AUC), and error rate. These all are related to the actual classification task. However, other performance-oriented indicators can also be considered, such as CPU time for 
training, CPU time for 
testing, and memory allocated to the classification model~\cite{ali2017accurate}.

Aside from the aforementioned challenges, there are also other sets of issues that are currently being investigated. These are:
\begin{itemize}
    \item Issues related to cross-domain, cross topic and/or cross-genre datasets;
    \item Issues related to the specificity of the used language;
    \item Issues regarding the style change of authors when the writing environment changes from offline to online;
    \item The balanced or imbalanced nature of datasets. 
\end{itemize}

Some examples which focus on these types of issues, alongside their solutions and/or findings, are presented next. 
 
Participants in the Identification Task at PAN-2018
~\cite{kestemont2018overview}, 
investigated two types of classifications. The corpus consists of fan-fiction texts written in English, French, Italian, Polish, and Spanish, and a set of questions and answers on several topics in English. 
First, they addressed the cross-domain AA, finding that heterogeneous ensembles of 
simple classifiers and compression models outperformed more sophisticated approaches based on deep learning.
Also, the set size is inversely correlated with attribution accuracy, especially for cases when more than 10 authors are considered. 
Second, they investigated the detection of style changes, where single-author and multi-author texts \change[E.E.]{are to be}{were
} distinguished. 
Techniques ranging from machine learning ensembles to deep learning with a rich set of features have been used to detect style changes, achieving the accuracy of up to nearly 90\% over the entire dataset and several reaching 100\%.  

The issue of cross-topic confusion is addressed in~\cite{altakrori-etal-2021-topic-confusion} for AA. This 
problem arises when the training topic differs from the test topic. In such a scenario, the types of errors caused by the topic can be distinguished from the errors caused by the detection of the writing style. The findings show that \change[E.E.]{considering parts of speech as features are least likely to be affected by topic variations}{classification is least likely to be affected by topic variations when parts of speech are considered as features}.


The analysis conducted in~\cite{sari2018topic} aimed to determine which approach, such as topic or style, is better for AA. The findings showed that online news, which have a wide variety of topics, are better classified using content-based features, while texts with less topical variation (e.g., legal judgments and movie reviews) benefit from using style-based features. 

In~\cite{sundararajan2018represents} it is shown that syntax (e.g., sentence structure) helps AA on cross-genre texts, while additional lexical information (e.g., parts of speech such as nouns, verbs, adjectives, and adverbs) helps to classify cross-topic and single-domain texts. It is also shown that syntax-only models may not be efficient.

Language-specific issues (e.g., the complexity and structure of \add[A.]{sentences}) are addressed in~\cite{al2020ensemble} in relation to the Arabic language. Ensemble methods were used to improve the effectiveness of the AA task. 

The authors of~\cite{CUSTODIO2021114866} propose solutions to address the many issues in AA (e.g., cross-domain, language specificity, writing environment) by introducing the concept of \emph{stacked classifiers}, which are built from words, characters, parts of speech n-grams, syntactic dependencies, word embeddings, and more. This solution proposes that these \emph{stacked classifiers} are dynamically included in the AA model according to the input. 

Two different AA approaches called ``writer-dependent'' and ``writer-independent'' were addressed in~\cite{pavelec2008using}. In the first approach, they used a Support Vector Machine (SVM) to build a model for each author. The second approach combined a feature-based description with the concept of dissimilarity to determine whether a text is written by a particular author or not, thereby reducing the problem to a two-class problem. The tests were performed on texts written in \change[A.]{the Portuguese language}{Portuguese}. For the first approach, 77 conjunctions and 94 adverbs were used to determine the authorship and the best accuracy 
results on the test set composed of 200 documents from 20 different authors were 83.2\%. 
For the second approach, the same set of documents and conjunctions was used, obtaining the best result of 75.1\% accuracy
. In~\cite{varela2010verbs}, along with conjunctions and adverbs, 50 verbs and 91 pronouns were added to improve the results obtained previously, achieving a 4\% improvement in both ``writer-dependent'' and ``writer-independent'' approaches.  

The challenges of \change[E.E.]{changes}{variations} in authors' style when the writing environment changes from traditional to online are addressed in~\cite{wang2021mode}. These investigations consider changes in sentence length, word usage, readability, and frequency use of some parts of speech. The findings show that shorter sentences and words, as well as more adjectives and pronouns, are used online.

The authors of~\cite{gonzalez2021new} proposed a feature extraction solution for AA. They investigated trigrams, bags of words, and most frequent terms in both balanced and imbalanced samples and showed with 79.68\% accuracy that an author's writing style can be \change[E.E.]{contained in}{identified by using} a single document.

\subsection{Comparison}
\label{related_work-comparison}

AA is a very important and currently intensively researched topic. However, the multitude of approaches makes it very difficult to have a unified view of the state-of-the-art results. In~\cite{tyo2022state}, authors highlight this challenge by noting significant differences in:
\begin{itemize}
  \item Datasets 
  \begin{itemize}
    \item In terms of size: small (CCAT50, CMCC, Guardian10), medium (IMDb62, Blogs50), and large (PAN20, Gutenberg);
    \item In terms of content: cross-topic ($\times_t$), cross-genre ($\times_g$), unique authors;
    \item In terms of imbalance (imb): i.e., standard deviation of the
number of documents per author; 
    \item In terms of topic confusion (as detailed in~\cite{altakrori-etal-2021-topic-confusion}).
  \end{itemize}
  \item Performance metrics
  \begin{itemize}
    \item In terms of type: accuracy, F1, c@1, recall, precision, macro-accuracy, AUC, R@8, and others;
    \item In terms of computation: even for the same performance metrics there were different ways of computing them.
  \end{itemize}
  \item Methods
  \begin{itemize}
    \item In terms of the feature extraction method,
    \begin{itemize}
      \item Feature-based: n-gram, summary statistics, co-occurrence graphs;
      \item Embedding-based: char embedding, word embedding, transformers
      \item Feature and embedding-based: BERT.
    \end{itemize}
  \end{itemize}
\end{itemize}

The work presented in~\cite{tyo2022state} tries to address and ``resolve'' these differences, bringing everything to a common denominator, even when that meant recreating some results. To differentiate between different methods, authors of~\cite{tyo2022state} grouped the results in 4 classes:
 \begin{itemize}
   \item Ngram: includes character n-grams, parts-of-speech and summary statistics as shown in~\cite{altakrori-etal-2021-topic-confusion, murauer-specht-2021-developing, bischoff2020importance, stamatatos2018masking};
   \item PPM: uses Prediction by Partial Matching (PPM) compression model to build a character-based model for each author, with works presented in~\cite{neal2017surveying, halvani2018cross};
   \item BERT: combines a BERT pre-trained language model with a dense layer for classification, as in~\cite{fabien-etal-2020-bertaa};
   \item pALM: the per-Author Language Model (pALM), also using BERT as described in~\cite{barlas2020cross}.
 \end{itemize}
 
 The results of the state of the art as presented in~\cite{tyo2022state} are shown in Table~\ref{tab:StateOfArt}. 
 \begin{table}[H]
\tablesize{\small}
\caption{State of the art \emph{macro-accuracy} of authorship attribution models. Information collected from~\cite{tyo2022state} (Tables \ref{tab:datasets} and \ref{tab:datasetDivSize}). 
\emph{Name} is the name of the dataset; \emph{No. docs} represents the number of documents in that dataset; \emph{No. auth} represents the number of authors; \emph{Content} indicates whether the documents are cross-topic ($\times_t$) or cross-genre ($\times_g$); \emph{W/D} stands for \emph{words per document}, representing the average length of documents; \emph{imb} represents the \emph{imbalance} of the dataset 
measured by the standard deviation of the number of documents per author.\label{tab:StateOfArt}}

\begin{adjustwidth}{-\extralength}{0cm}
\setlength{\cellWidtha}{\fulllength/10-2\tabcolsep+0.2in}
\setlength{\cellWidthb}{\fulllength/10-2\tabcolsep-0in}
\setlength{\cellWidthc}{\fulllength/10-2\tabcolsep-0in}
\setlength{\cellWidthd}{\fulllength/10-2\tabcolsep-0in}
\setlength{\cellWidthe}{\fulllength/10-2\tabcolsep-0in}
\setlength{\cellWidthf}{\fulllength/10-2\tabcolsep-0in}
\setlength{\cellWidthg}{\fulllength/10-2\tabcolsep-0in}
\setlength{\cellWidthh}{\fulllength/10-2\tabcolsep-0.2in}
\setlength{\cellWidthi}{\fulllength/10-2\tabcolsep-0in}
\setlength{\cellWidthj}{\fulllength/10-2\tabcolsep-0in}
\scalebox{1}[1]{\begin{tabularx}{\fulllength}{>{\centering\arraybackslash}m{\cellWidtha}>{\centering\arraybackslash}m{\cellWidthb}>{\centering\arraybackslash}m{\cellWidthc}>{\centering\arraybackslash}m{\cellWidthd}>{\centering\arraybackslash}m{\cellWidthe}>{\centering\arraybackslash}m{\cellWidthf}>{\centering\arraybackslash}m{\cellWidthg}>{\centering\arraybackslash}m{\cellWidthh}>{\centering\arraybackslash}m{\cellWidthi}>{\centering\arraybackslash}m{\cellWidthj}}
\toprule

\multicolumn{6}{c}{\textbf{Dataset} } & \multicolumn{4}{c}{\emph{\textbf{Macro-Accuracy}} \textbf{(\%) for Investigation Type} }\\\midrule
    \textbf{Name} & \textbf{No. Docs} & \textbf{No. Auth} & \textbf{Content}& \textbf{W/D}& \textbf{Imb}& \textbf{Ngram} & \textbf{PPM} & \textbf{BERT} & \textbf{pALM}\\\midrule
    CCAT50 & 5000&50& - & 506& 0& 76.68& 69.36 & 65.72 & 63.36\\
    CMCC & 756&21& $\times_t$ $\times_g$&601& 0& 86.51 & 62.30 & 60.32 & 54.76\\
    Guardian10& 444&13 & $\times_t$ $\times_g$& 1052& 6.7& 100 & 86.28 & 84.23 & 66.67 \\
    IMDb62 & 62,000&62 & -&349& 2.6& 98.81 & 95.90 & 98.80 & -\\
    Blogs50 & 66,000&50&- &122& 553& 72.28& 72.16 & 74.95 & - \\
    PAN20 & 443,000&278,000 &$\times_t$ &3922&2.3& 43.52 & - & 23.83 & -\\
    Gutenburg & 29,000&4500& -&66,350 &10.5& 57.69& - & 59.11 & - \\

\bottomrule
\end{tabularx}}
\end{adjustwidth}

\end{table}

As can be seen in Table~\ref{tab:StateOfArt}, the methods in the Ngram 
class generally work best. However, BERT-class methods can perform better on large training sets\change[A.]{, but not always (e.g., for PAN20 dataset)}{ that are not cross-topic and/or cross-genre}. The authors of~\cite{tyo2022state} state that from their investigations it can be inferred that Ngram-class methods are preferred for datasets that have less than 50,000 words per author, while BERT-class methods should be preferred for datasets with over 100,000 words per author.

\section{Proposed Dataset}
\label{proposed_dataset}

The texts considered are Romanian stories, short stories, fairy tales, novels, articles, and sketches. 

There are 400 such texts of different lengths, ranging from 91 to 39,195 words. 
\remove[A.]{A visual representation of the average (and standard deviation)} \add[A.]{Table}~\ref{tab:datasetDivSize} \add[A.]{presents the averages and standard deviations} of the number of words, unique words, and the ratio of words to unique words for each author. \remove[A.]{is shown in Figure~2. As can be seen, t}There are differences up to 
\change[A.]{5000}{almost 7000} words between the average word counts (e.g., between Slavici and Oltean). 
For unique words, the difference between averages goes up to more than 1300 
unique words (e.g., between Eminescu and Oltean)
. Even the ratio of total words to unique words is a significant difference between the authors (e.g., between Slavici and Oltean)
.

\begin{table}[H]
\caption{Diversity of the considered dataset in terms of the length of the texts (i.e., number of words). 
\emph{Author} is the author's name (the last name is in bold); 
\emph{Average} is the mean number of words per text written by the corresponding author; \emph{StdDev} is the standard deviation; \emph{Average-Unique} is the mean number of unique words; \emph{StdDev-Unique} is the standard deviation on unique words; \emph{Average-Ratio} is the mean number of the ratio of total words to unique words; \emph{StdDev-Ratio} is the standard deviation of the ratio of total words to unique words. \label{tab:datasetDivSize}}
  \begin{adjustwidth}{-\extralength}{0cm}
    \begin{tabularx}{\fulllength}{l C C C C C C}
      \toprule
      \textbf{Author} & \textbf{Average} & \textbf{StdDev} & \textbf{Average-Unique} & \textbf{StdDev-Unique} & \textbf{Average-Ratio} & \textbf{StdDev-Ratio}\\\midrule
      Ion \textbf{Creangă} 
      & 3679.34 & 3633.42 & 1061.90 & 719.38 & 3.01 & 0.94\\
      Barbu Ştefănescu \textbf{Delavrancea} & 4166.39 & 3702.33 & 1421.34 & 948.41 & 2.66 & 0.58\\
      Mihai \textbf{Eminescu} & 5854.52 & 7858.89 & 1656.96 & 1716.08 & 2.92 & 0.87\\
      Nicolae \textbf{Filimon} & 2734.32 & 2589.72 & 1040.09 & 729.81 & 2.42 & 0.50\\
      Emil \textbf{Gârleanu} & 843.05 & 721.06 & 411.19 & 234.71 & 1.88 & 0.32\\
      Petre \textbf{Ispirescu} & 3302.80 & 1531.36 & 1017.73 & 340.37 & 3.10 & 0.49\\
      Mihai \textbf{Oltean} & 553.75 & 484.00 & 282.56 & 201.18 & 1.79 & 0.31\\
      Emilia \textbf{Plugaru} & 2253.88 & 2667.38 & 756.70 & 581.88 & 2.54 & 0.64\\
      Liviu \textbf{Rebreanu} & 2284.12 & 1971.88 & 889.70 & 550.92 & 2.36 & 0.44\\
      Ioan \textbf{Slavici} & 7531.54 & 8969.77 & 1520.42 & 1041.40 & 3.96 & 1.62\\
      \bottomrule
    \end{tabularx}
  \end{adjustwidth}
\end{table}

\remove[A.]{In Figures~2a and 2b,} Eminescu and Slavici, the two authors with the \change[A.]{first two}{largest} averages also have large standard deviations for the number of words and the number of unique words. This means that \change[E.E.]{they have both very short or very long texts}{their texts range from 
very short to 
very long}. Gârleanu and Oltean have the shortest texts, as their average number of words and unique words and the corresponding standard deviations are the smallest. 

There is also a correlation between the three \remove[A.]{ graphs from Figure~2 and this} \add[A.]{groups of values (pertaining to the words, unique words, and the ratio between the two) that} is to be expected as a larger or smaller number of words would contain a similar proportion of unique words or the ratio of the two\change[A.]{. However,}{, while} the standard deviations of the ratio of total words to
unique words tend to be more similar.\remove[A.]{, as can be seen in Figure~2c.} However, Slavici has a very high ratio, which means that there are texts in which he repeats the same words more often, and in other texts, he does not. There is also a difference between Slavici and Eminescu here because even if they have similar word count average and unique word count average, their ratio \remove[A.]{in Figure~2c}differs. \remove[A.]{In Figure~2c,}Eminescu has a similar representation in terms of ratio and standard deviation with his lifelong friend Creangă, which can mean that both may have similar tendencies in \change[E.E.]{word reuse}{reusing words}.

 \change[A.]{Figure~3}{Table}~\ref{tab:datasetDivFeatSize} shows the averages of the number of features that are contained in the texts corresponding to each author. The pattern depicted \remove[A.]{in the graphics}here is similar to that in \change[A.]{Figure~2}{Table}~\ref{tab:datasetDivSize}, which is to be expected. However, standard deviations tend to be similar for all authors. These standard deviations are considerable in size, being on average as follows:
 \begin{itemize}
   \item 4.16 on the set of 56 features (i.e., the list of prepositions), 
   \item 23.88 on the set of 415 features (i.e., the list of prepositions and adverbs), 
   \item 25.38 on the set of 432 features (i.e., the list of prepositions, adverbs, and conjunctions). 
 \end{itemize}

 This means that \change[E.E.]{even for the same author, the occurrence of words in the considered feature lists differs within the considered set of texts}{the frequency of feature occurrence differs even in the texts written by the same author}. 

\begin{table}[H]
\caption{Diversity of the considered dataset in terms of the number of occurrences of the considered features in the texts. \emph{Author} is the author's name (the last name is in bold); 
\emph{Average-P} is the average number of the occurrence of the considered prepositions in the texts corresponding to each author; \emph{StdDev-P} is the standard deviation for the occurrence of the prepositions; \emph{Average-PA} is the average number of the occurrence of the considered prepositions and adverbs; \emph{StdDev-PA} is the standard deviation of the number of the occurrence of the considered prepositions and adverbs; \emph{Average-PAC} is the average number of the occurrence of the considered prepositions, adverbs, and conjunctions; \emph{StdDev-PAC} is the standard deviation of the number of the occurrence of the considered prepositions, adverbs, and conjunctions. \label{tab:datasetDivFeatSize}}
  \begin{adjustwidth}{-\extralength}{0cm}
    \begin{tabularx}{\fulllength}{l C C C C C C}
      \toprule
      \textbf{Author} & \textbf{Average-P} & \textbf{StdDev-P} & \textbf{Average-PA} & \textbf{StdDev-PA} & \textbf{Average-PAC} & \textbf{StdDev-PAC}\\\midrule
      Ion \textbf{Creangă} & 19.90 & 4.94 & 79.21 & 30.11 & 88.34 & 31.86 \\
      Barbu Ştefănescu \textbf{Delavrancea} & 19.14 & 3.67 & 73.43 & 27.79 & 81.82 & 29.81 \\
      Mihai \textbf{Eminescu} & 21.85 & 7.18 & 80.04 & 34.11 & 90.04 & 36.22 \\
      Nicolae \textbf{Filimon} & 18.26 & 3.52 & 61.94 & 18.12 & 70.50 & 19.25 \\
      Emil \textbf{Gârleanu} & 14.65 & 3.01 & 48.12 & 16.11 & 53.21 & 17.19 \\
      Petre \textbf{Ispirescu} & 19.93 & 3.14 & 79.60 & 17.32 & 89.63 & 18.52 \\
      Mihai \textbf{Oltean} & 11.88 & 3.82 & 33.16 & 17.51 & 37.69 & 18.96 \\
      Emilia \textbf{Plugaru} & 16.13 & 3.61 & 69.83 & 22.62 & 77.48 & 23.58 \\
      Liviu \textbf{Rebreanu} & 17.25 & 4.07 & 73.88 & 25.65 & 82.62 & 27.37 \\
      Ioan \textbf{Slavici} & 21.29 & 4.72 & 96.08 & 29.48 & 105.87 & 31.09 \\
      \bottomrule
    \end{tabularx}
  \end{adjustwidth}
\end{table}

The considered texts are collected from 4 websites and are written by 10 different authors, as shown in Table~\ref{tab:textStats}. \add[A.]{The diversity of sources is relevant from a twofold perspective. First, especially for old texts, it is difficult to find or determine which is the original version. Second, there may be differences between versions of the same text either because some words are no longer used or have changed their meaning, or because fragments of the text may be added or subtracted.} \change[E.E.]{As can be observed, it can happen that for an author the sources are 2 or even 3 different sites.}{For some authors, texts are sourced from multiple websites.} 

\begin{table}[H]
    \caption{List of authors (the author's last name is in bold), 
    the number of texts considered for each author (total number is in bold)
    , and their source (i.e., the website from which they were collected).\label{tab:textStats}}
    \begin{adjustwidth}{-\extralength}{0cm}
    \begin{tabularx}{\fulllength}{l C C C C C}
        \toprule
        \textbf{Author}& \textbf{No. of Texts} & \href{https://www.povesti.org}{povesti.org} & \href{https://povesti-ro.weebly.com/}{povesti-ro.weebly.com} & \href{https://ro.wikisource.org/wiki/}{ro.wikisource.org} & \href{ https://www.povesti-pentru-copii.com/}{povesti-pentru-copii.com}\\ \midrule
        Ion \textbf{Creangă}& \textbf{28} &&&4&24\\
        Barbu Ştefănescu \textbf{Delavrancea}& \textbf{44}&&2&28&14\\
        Mihai \textbf{Eminescu}& \textbf{27} &&&21&6\\
        Nicolae \textbf{Filimon}& \textbf{34}&&&31&3\\
        Emil \textbf{Gârleanu}& \textbf{43}&&34&9&\\
        Petre \textbf{Ispirescu}& \textbf{40}&&2&1&37\\
        Mihai \textbf{Oltean}& \textbf{32}&32&&&\\
        Emilia \textbf{Plugaru}& \textbf{40}&&&&40\\
        Liviu \textbf{Rebreanu}& \textbf{60}&&&60&\\
        Ioan \textbf{Slavici}& \textbf{52}&&3&39&10\\\midrule
        TOTAL & \textbf{400}	&32 &41 & 193&134 \\
	\bottomrule
    \end{tabularx}
    \end{adjustwidth}
\end{table}

\change[E.E.]{This diversity is also intended}{The diversity of the texts is intentional} because we wanted to emulate a more likely scenario where all these characteristics might not be controlled. This is because, for future texts to be tested on the trained models, the text length
, the source, and the type of writing cannot be controlled or imposed. 

To highlight the differences between the time frames of the periods in which the authors lived and 
wrote the considered texts, as well as the environment 
from which the texts were intended to be 
read, we gathered the information presented in Table~\ref{tab:textStats2}. It can be seen that the considered texts were written in \change[E.E.]{a time span spanning}{the time span of} three centuries. This also brings an increased diversity between texts, since within such a large time span there have been significant developments in terms of language (e.g., diachronic developments), writing style relating to the desired reading medium (e.g., paper or online), topics (e.g., general concerns and concerns that relate to a particular time), and viewpoints (e.g., a particular worldview).

\begin{table}[H]
    \caption{List of authors, time spans of the periods in which the authors lived and wrote the considered texts and the medium from which the readers read their texts. \emph{Author} is the author's name (the last name is in bold); 
    \emph{Life} is the lifetime of the author; \emph{Publication} is the publication interval of the texts (note: the information presented here was not always easily accessible and some sources would contradict in terms of specific years, however, this information should be considered more as an indicative coordinate and should not be taken literally, the goal being that the literary texts be temporally framed in order to have a perspective on the period in which they were written/published); \emph{Century} is a coarser temporal framing of the periods in which the texts were written; \emph{Medium} is the environment 
    from which most of the readers read the author's texts. \label{tab:textStats2}}
    \begin{adjustwidth}{-\extralength}{0cm}
    \begin{tabularx}{\fulllength}{C l C C C C}
    \toprule
        \#&\textbf{Author}& \textbf{Life} & \textbf{Publication} &\textbf{Century} &\textbf{Medium}\\ 
        \midrule
        0& Ion \textbf{Creangă}&1837--1889&1874--1898& 19th  &paper\\
        1& Barbu Ştefănescu \textbf{Delavrancea}&1858--1918&1884--1909& 19{th}--20{th}  &paper\\
        2& Mihai \textbf{Eminescu}& 1850--1889&1872--1865& 19{th}  &paper\\
        3& Nicolae \textbf{Filimon}& 1819--1865&1857--1863& 19{th} &paper\\
        4& Emil \textbf{Gârleanu}&1878--1914&1907--1915& 20{th} &paper\\
        5& Petre \textbf{Ispirescu}&1830--1887& 1882--1883& 19{th} &paper\\
        6& Mihai \textbf{Oltean}&1976--& 2010--2022& 21{th} &paper and online\\
        7& Emilia \textbf{Plugaru}&1951--&2010--2017& 21{th} &paper and online\\
        8& Liviu \textbf{Rebreanu}&1885--1944&1908--1935& 20{th} &paper\\
        9& Ioan \textbf{Slavici}&1848--1925&1872--1920& 19{th}--20{th} &paper\\
        \bottomrule
    \end{tabularx}
    \end{adjustwidth}
\end{table}

The diversity of the texts also pertains to the type of writing, i.e., stories, short stories, fairy tales, novels, articles, and sketches. Table~\ref{tab:textStats3} shows the distribution of these types of writing among the texts belonging to the 10 authors. The difference in the type of writing has an impact on the length of the texts (for example, a \emph{novel} is considerably longer than a \emph{short story}), genre (for example, \emph{fairy tales} have more allegorical worlds that can require a specific style of writing), the topic (for example, an \emph{article} may describe more mundane topics, requiring a different type of discourse compared to the other types of writing). 

\begin{table}[H]
    \caption{List of authors 
    and types of writing of the considered texts. \emph{Author} is the author's name (the last name is in bold); 
    \emph{Article $^*$} include, in addition to articles written for various newspapers and magazines, other types of writing that did not fit into the other categories, but relate to this category, such as \emph{prose}, \emph{essays}, and theatrical or musical \emph{chronicles}. Total number of texts per type are in bold. 
    \label{tab:textStats3}}
    \begin{adjustwidth}{-\extralength}{0cm}
    \begin{tabularx}{\fulllength}{C l C C C C C C}
        \toprule
        \textbf{\#}&\textbf{Author}& \textbf{Novel} & \textbf{Story} & \textbf{Short Story} & \textbf{Fairy Tale }& \textbf{\emph{Article} *} & \textbf{Sketch} \\ \midrule
        0& Ion \textbf{Creangă} & 5 & 12 & 11 & & & \\
        1& Barbu Ştefănescu \textbf{Delavrancea} & & & 37&7 & & \\
        2& Mihai \textbf{Eminescu} & 1& 1& 4& 7& 14& \\
        3& Nicolae \textbf{Filimon} & 6& & 5& 3& 20& \\
        4& Emil \textbf{Gârleanu} & & & 43& & & \\
        5& Petre \textbf{Ispirescu} & & 1& 1& 38 & &\\
        6& Mihai \textbf{Oltean} & & & 32& & &\\
        7& Emilia \textbf{Plugaru} & & 40& & & &\\
        8& Liviu \textbf{Rebreanu} & & 46& & & & 14\\
        9& Ioan \textbf{Slavici} & & 14& 38 & & \\\midrule
        & TOTAL & \textbf{12}	& \textbf{113}& \textbf{171} & \textbf{55} & \textbf{35} & \textbf{14} \\
        \bottomrule
    \end{tabularx}
    \end{adjustwidth}
\end{table}
  
Regarding the list of possible features, we selected as elements to identify the author of a text \emph{inflexible parts of speech} (IPoS) (i.e., those that do not change their form in the context of communication): conjunctions, prepositions, interjections, and adverbs. Of these, we only considered those that were single-word and we removed the words that may represent other parts of speech, as some of them may have different functions depending on the context, and we did not use any syntactic or semantic processing of the text 
to carry out such an investigation. 

We collected a list of 24 conjunctions that we checked on \href{https://dexonline.ro/}{dexonline.ro} (i.e.,  site that contains explanatory dictionaries of the Romanian language) 
not to be any other part of speech (not even among the inflexible ones). We also considered 3 short forms, thus arriving at a list of 27 conjunctions. The process of selecting prepositions was similar to that of selecting conjunctions, resulting in a list of 85 (including some short forms). 

The lists of interjections and adverbs were taken from:
\begin{itemize}
  \item \href{https://ro.wiktionary.org/wiki/Categorie:Interjec%C8%9Bii_%C3%AEn_rom%C3%A2n%C4%83
  }{ro.wiktionary.org/wiki/Categorie:Interjecții\_în\_română}, accessed on 20 October 2022
  \item \href{https://ro.wiktionary.org/wiki/Categorie:Adverbe_%C3%AEn_rom%C3%A2n%C4%83
  }{ro.wiktionary.org/wiki/Categorie:Adverbe\_în\_română}, accessed on 20 October 2022
\end{itemize}

To compile the lists of interjections and adverbs, we again considered only single-word ones and we eliminated words that may represent other parts of speech (e.g., proper nouns, nouns, adjectives, verbs), resulting in lists of 290 interjections and 670 adverbs. 

\add[A.]{The lists of the aforementioned IPoS also contain archaic forms in order to better identify the author. This is an important aspect that has to be taken into consideration (especially for our dataset which contains texts that were written over a time span of 3~centuries), as language is something that evolves and some words change as form and sometimes even as meaning or the way they are used.}

From the lists corresponding to the considered IPoS features
, we use only those that appear in the texts. Therefore, the actual lists of prepositions, adverbs, and conjunctions may be shorter. Details of the texts and the lists of inflexible parts of speech used can be found at reference~\cite{sanda2022kaggle}.

\section{Compared Methods}
\label{compared_methods}
Below we present the methods we will use in our investigations.
\subsection{Artificial Neural Networks}
 Artificial neural networks (ANN) is a machine learning method that applies the principle function approximation through learning by example (or based on provided training information)~\cite{zurada1992introduction}. An ANN contains artificial neurons (or processing elements), organized in layers and connected by weighted arcs. The learning process takes place by adjusting the weights during the training process so that based on the input dataset the output outcome is obtained. Initially, these weights are chosen randomly. 
 
 The artificial neural structure is feedforward and has at least three layers: input, hidden (one or more), and output.
 
The experiments in this paper were performed using fast artificial neural network (FANN)~\cite{steffen2005neural} library. The error is RMSE. For the test set, the number of incorrectly classified items is also calculated.

\subsection{Multi-Expression Programming}
Multi-expression programming (MEP) is an evolutionary algorithm for generating computer programs. It can be applied to symbolic regression, time-series, and classification problems~\cite{oltean_mep_class}. It is inspired by genetic programming~\cite{koza} and uses three-address code~\cite{aho86} for the representation of programs. 

MEP experiments use the MEPX software~\cite{MEPX}.

\subsection{K-Nearest Neighbors}
K-nearest neighbors (k-NN)~\cite{fix1951discriminatory,fix1952discriminatory, altman1992introduction} is a simple classification method based on the concept of instance-based learning~\cite{aha1991instance}. It finds the $k$ items, in the training set, that are closest to the test item and assigns the latter to the class that is most prevalent among these $k$ items found.

The source code of k-NN used in this paper is written by us \add[A.]{and is available at} \href{https://github.com/sanda-avram/ROST-source-code}{github.com/sanda-avram/ROST-source-code}, (accessed on 29 November 2022) along other scripts and programs we wrote to perform the tests.

\subsection{Support Vector Machine}
A support vector machine (SVM)~\cite{boser1992training} is also a classification principle based on machine learning with the maximization (support) of separating distance/margin (vector). As in k-NN, SVM represents the items as points in a high-dimensional space and tries to separate them using a hyperplane. The particularity of SVM lies in the way in which such a hyperplane is selected, i.e., selecting the hyperplane that has the maximum distance to any item. 

LIBSVM~\cite{hsu2003practical,chang2011libsvm} is the support vector machine library that we used in our experiments. It supports classification, regression, and distribution estimation. 

\subsection{Decision Trees with C5.0}
Classification can \textls[-15]{be completed by representing the acquired knowledge as decision trees~\cite{quinlan1986induction}}. A decision tree is a directed graph in which all nodes (except the root) have exactly one incoming edge. The root node has no incoming edge. All nodes that have outgoing edges are called internal (or test) nodes. All other nodes are called leaves (or decision) nodes. Such trees are built starting from the root by top--down inductive inference based on the values of the items in the training set. So, within each internal node, the instance space is divided into two or more sub-spaces based on the input attribute values. An internal node may consider a single attribute. Each leaf is assigned to a class. Instances are classified by running them through the tree starting from the root to the leaves. 

See5 and C5.0~\cite{RuleQuest} are data mining tools that produce classifiers expressed as either decision trees or rulesets, which we have used in our experiments.

\section{Numerical Experiments}
\label{numerical_experiments}
To prepare the dataset for the actual building of the classification model, the texts in the dataset were shuffled and divided into training (50\%), validation (25\%), and test (25\%) sets, as detailed in Table~\ref{tab:textDiv}. In cases where we only needed training and test sets, we concatenated the validation set to the training set. We reiterated the process 
(i.e., shuffle and split 50\%--25\%--25\%) three times and, thus, obtained three different training--validation--test shuffles from the considered dataset. 

\begin{table}[H]
    \caption{List of authors (the author's last name is in bold); 
    the number of texts and their distribution on the training, validation, and test sets. The total number of texts per author, per set, and grand total are in bold. 
    \label{tab:textDiv}}
    \begin{adjustwidth}{-\extralength}{0cm}
    \begin{tabularx}{\fulllength}{c l C C C C}
        \toprule
        \#&\textbf{Author}&\textbf{No. of Texts}& \textbf{TrainSet Size}& \textbf{ValidationSet Size}& \textbf{TestSet Size}\\ \midrule
        0& Ion \textbf{Creangă}& \textbf{28}& 14& 7& 7\\
        1& Barbu Ştefănescu \textbf{Delavrancea}& \textbf{44}& 22& 11& 11\\
        2& Mihai \textbf{Eminescu}& \textbf{27}& 15& 6& 6\\
        3& Nicolae \textbf{Filimon}& \textbf{34}& 18& 8& 8\\
        4& Emil \textbf{Gârleanu}& \textbf{43}& 23& 10& 10\\
        5& Petre \textbf{Ispirescu}& \textbf{40}& 20& 10& 10\\
        6& Mihai \textbf{Oltean}& \textbf{32}& 16& 8& 8\\
        7& Emilia \textbf{Plugaru}& \textbf{40}& 20& 10& 10\\
        8& Liviu \textbf{Rebreanu}& \textbf{60}& 30& 15& 15\\
        9& Ioan \textbf{Slavici}& \textbf{52}& 26& 13& 13\\\midrule
        & TOTAL & \textbf{400}	&\textbf{204}&	\textbf{98}&	\textbf{98}\\
        \bottomrule
    \end{tabularx}
    \end{adjustwidth}
\end{table}

Before building a numerical representation of the dataset as vectors of the frequency of occurrence of the considered features, we made a preliminary analysis to determine which of the inflexible parts of speech are more prevalent in our texts. Therefore, we counted the number of occurrences of each of them based on the lists described in Section~\ref{proposed_dataset}. The findings are detailed in Table~\ref{tab:IPoS_stats}. 

\begin{table}[H]
    \caption{The occurrence of inflexible parts of speech considered. \emph{IPoS} stands for \emph{Inflexible part of speech}; \emph{No. of occurrence} \change[A.]{represents the total number of all words in a corresponding list of IPoS occurring in the texts}{is the total number of occurrences of the considered IPoS in all texts}; \emph{\% from total words} represents the percentage corresponding to the \emph{No. of occurrence} in terms of the total number of words in all texts (i.e., 1,342,133); \emph{No. of files} represents the number of texts in which at least one word from the corresponding IPoS list appears; \emph{Avg. per file} represents the \emph{No. of occurrence} divided by the total number of texts/files (i.e., 400); and \emph{No. of IPoS} represents the list length (i.e., the number of words) for each corresponding IPoS.\label{tab:IPoS_stats}}
    \begin{adjustwidth}{-\extralength}{0cm}
    \begin{tabularx}{\fulllength}{C C C C C C}
        \toprule
        \textbf{IPoS} & \textbf{No. of Occurrence} & \textbf{\% from Total Words} & \textbf{No. of Files} & \textbf{Avg. per File} & \textbf{No. of IPoS}\\ \midrule 
        conjunctions & 119,568 & 8.90 & 400 & 298.92 & 27\\
        prepositions & 176,733 & 13.16 & 400 & 441.83 & 85\\ 
        interjections & 6614 & 0.49 & 356 & 16.53 & 290\\ 
        adverbs & 127,811 & 9.52 & 400 & 319.52 & 670\\
         \bottomrule
    \end{tabularx}
    \end{adjustwidth}
\end{table}

Based on the data presented here, we decided not to consider interjections because they do not appear in all files (i.e., 44 files do not contain any interjections), and in the other files, their occurrence is much less compared to the rest of the IPoS considered. This investigation also allowed us to decide the order in which these IPoS will be considered in our tests. Thus, the order of investigation is prepositions, adverbs, and conjunctions.

Therefore, we would first consider only prepositions, then add adverbs to this list, and finally add conjunctions as well. The process of shuffling and splitting the texts into training--validation--test sets (described at the beginning of the current section, i.e., Section~\ref{numerical_experiments}) was reiterated once more for each feature list considered. We, therefore, obtained different dataset representations, which we will refer further as described in Table~\ref{tab:datasetsNames}. The last 3~entries (i.e., ROST-PC-1, ROST-PC-2, and ROST-PC-3) were used in a single experiment.
 
\begin{table}[H]
    \centering
    \caption{Names used in the rest of the paper refer to the different dataset representations and their shuffles. Only the first 9 entries (with the \emph{Designation} written in bold) 
    were used for the entire set of investigations. \label{tab:datasetsNames}}
    \begin{tabularx}{\textwidth}{C l l C}
        \toprule
        \textbf{\#} & \textbf{Designation} & \textbf{Features to Represent the Dataset} & \textbf{Shuffle}\\\midrule
        1 & \textbf{ROST-P-1} & prepositions &  \#1 \\
        2 & \textbf{ROST-P-2} & prepositions& \#2 \\
        3 & \textbf{ROST-P-3} & prepositions& \#3 \\\midrule
        4 & \textbf{ROST-PA-1} & prepositions and adverbs & \#1 \\
        5 & \textbf{ROST-PA-2} &  prepositions and adverbs & \#2 \\
        6 & \textbf{ROST-PA-3} &  prepositions and adverbs & \#3 \\\midrule
        7 & \textbf{ROST-PAC-1} &  prepositions, adverbs and conjunctions & \#1 \\
        8 & \textbf{ROST-PAC-2} &  prepositions, adverbs and conjunctions & \#2 \\
        9 & \textbf{ROST-PAC-3} &  prepositions, adverbs and conjunctions & \#3 \\\midrule
        10 & ROST-PC-1 &  prepositions and conjunctions & \#1 \\
        11 & ROST-PC-2 &  prepositions and conjunctions & \#2 \\
        12 & ROST-PC-3 &  prepositions and conjunctions & \#3 \\
        \bottomrule
    \end{tabularx}
\end{table}

Correspondingly, we created different representations of the dataset as vectors of the frequency of occurrence of the considered feature lists. All these representations (i.e., training-validation-test sets) can be found as text files at reference~\cite{sanda2022kaggle}. These files contain feature-based numerical value representations for a different text on each line. On the last column of 
these files, are numbers from 0 to 9 corresponding to the author, as specified in the first columns of Tables~\ref{tab:textStats2}
--\ref{tab:textDiv}.

\subsection{Results}
\add[A.]{The parameter setting for all 5 methods are presented in Appendix~}\ref{app:param-settings}, while \linebreak  Appendix~\ref{app:prereq-results} \add[A.]{contains some prerequisite tests.}

Most results are presented in a tabular format. The percentages contained in the cells under the columns named \emph{Best}, \emph{Avg}, or \emph{Error} may be highlighted using bold text or gray background. In these cases, the percentages in bold represent the best individual results (i.e., obtained by the respective method on any ROST-*-* in the dataset, out of the 9 \add[A.]{representations} mentioned above), while the gray-colored cells contain the best overall results (i.e., compared to all methods on that specific ROST-X-n representation of 
the dataset).

\subsubsection{ANN}
 \add[A.]{Results that showed that ANN is a good candidate to solve this kind of problem and prerequisite tests that determined the best ANN configuration (i.e., number of neurons on the hidden layer) for each dataset representation are detailed in Appendix~}\ref{app:prereq-results-ANN}.
The best values obtained for test errors and the number of neurons on the hidden layer for which these ``bests'' occurred are given in Table~\ref{tab:rez-ANN}. These results show that the best test error rates were mainly generated by ANNs that have a number of neurons between 27 and 49. The best test error rate obtained with this method was $23.46\%$ for ROST-PAC-3, while the best average was $36.93\%$ for ROST-PAC-2. 

\begin{table}[H]
    \caption{ANN results 
    on the considered datasets. On each set, 30 runs are performed by ANNs with the hidden layer containing from 5 to 50 neurons. The number of incorrectly classified data is given as a percentage (the best results obtained by ANN on any ROST-*-* dataset representation are in bold)
    . \textit{Best} stands for the best solution (out of 30 runs on each of the 46 ANNs), \textit{Avg} stands for \textit{Average} (over 30 runs), \textit{StdDev} stands for \textit{Standard Deviation}, and \textit{No. of neurons} stands for the number of neurons in the hidden layer of the ANN that produced the best solution. The best result obtained by ANN compared to all methods for a given ROST-X-n dataset representation is in a gray cell. 
    \label{tab:rez-ANN}}
    \begin{tabularx}{\textwidth}{CCCCC}
        \toprule
        \textbf{Dataset} & \textbf{Best} & \textbf{Avg} & \textbf{StdDev} & \textbf{No. of Neurons}\\ \midrule
        ROST-P-1 & 61.22\% & 76.70\%& 6.30 & 46 \\ 
        ROST-P-2 & 60.20\% & 80.27\%& 10.58 & 36 \\ 
        ROST-P-3 & 57.14\% & 80.95\%& 10.30 & 28 \\ \hline
        ROST-PA-1 & \cellcolor{gray!25}24.48\% & 45.03\%& 8.15 & 40 \\ 
        ROST-PA-2 & 24.48\% & 41.73\%& 5.78 & 45 \\ 
        ROST-PA-3 & \cellcolor{gray!25}26.53\% & 47.82\%& 9.82 & 27 \\ \hline
        ROST-PAC-1 & 24.48\% & 38.16\%& 5.11 & 49 \\ 
        ROST-PAC-2 & \cellcolor{gray!25}24.48\% & \textbf{36.93\%}& 4.80 & 40 \\ 
        ROST-PAC-3 & \cellcolor{gray!25}\textbf{23.46\%} & 37.21\%& 4.96 & 41\\  
  \noalign{\hrule height 1pt} 
\end{tabularx}
\end{table}

\subsubsection{MEP}
\add[A.]{Results that showed that MEP can handle this type of problem are described in Appendix~}\ref{app:prereq-results-MEP}.

We are interested in the generalization ability of the method. For this purpose, we performed full (30) runs on all datasets. The results, on the test sets, are given in Table~\ref{tab:rez-MEP}. 
\begin{table}[H]
    \caption{MEP results 
    on the considered datasets. A total of 30 runs are performed. The number of incorrectly classified data is given as a  percentage (the best results obtained by MEP on any ROST-*-* dataset representation are in bold)
    . \textit{Best} stands for the best solution (out of 30 runs), \textit{Avg} stands for \textit{Average} (over 30 runs) and \textit{StdDev} stands for \textit{Standard Deviation}. The best result obtained by MEP compared to all methods for a given ROST-X-n dataset representation is in a gray cell. 
    \label{tab:rez-MEP}}
    \begin{tabularx}{\textwidth}{CCCC}
        \toprule
        \textbf{Dataset} & \textbf{Best} & \textbf{Avg} & \textbf{StdDev} \\ \midrule
        ROST-P-1 & 54.08\%& 61.32\%& 4.11\\
        ROST-P-2 & 52.04\%& 62.51\%& 4.46\\
        ROST-P-3 &48.97\%& 58.84\%& 4.16\\ \hline
        ROST-PA-1 &29.59\%& 36.49\%& 4.52\\
        ROST-PA-2 &\cellcolor{gray!25} \textbf{20.40\%}& \textbf{27.95\%}& 3.87 \\
        ROST-PA-3 &29.59\%& 39.93\%& 4.53\\ \hline
        ROST-PAC-1 & 27.55\%& 33.84\% & 2.86\\
        ROST-PAC-2 & 26.53\%& 34.89\% & 4.58\\
        ROST-PAC-3 &\cellcolor{gray!25} 23.46\%& 34.38\% & 4.54\\
        \noalign{\hrule height 1pt} 
    \end{tabularx}
\end{table}

With this method, we obtained an overall ``best'' on all ROST-*-*, which is $20.40\%$, and also an overall ``average'' best with a value of $27.95\%$, both for ROST-PA-2.

One big problem is overfitting. The error on the training set is low (they are not given here, but sometimes are below 10\%). However, on the validation and test sets the errors are much higher (2 or 3 times higher). This means that the model suffers from overfitting and has poor generalization ability. This is a known problem in machine learning and is usually corrected by providing more data (for instance more texts for an author).


\subsubsection{k-NN}
\add[A.]{Preliminary tests and their results for determining the best value of $k$ for each dataset representation are presented in Appendix~}\ref{app:prereq-results-kNN}.

The best k-NN results are given in Table~\ref{tab:rez-kNN} with the corresponding value of $k$ for which these ``bests'' were obtained. It can be seen that for all ROST-P-*, the values of $k$ were higher (i.e., $k\ge8$) than those for ROST-PA-* or ROST-PAC-* (i.e., $k\le4$). The best value obtained by this method was $29.59\%$ for ROST-PAC-2 and ROST-PAC-3.

\begin{table}[H]
  \caption{k-NN results 
 on the considered datasets. In total, 30 runs are performed with $k$ varying with the run index. The number of incorrectly classified data is given as a percentage (the best results obtained by k-MM on any ROST-*-* dataset representation are in bold)
 . \textit{Best} stands for the best solution (out of the 30 runs), \textit{k} stands for the value of $k$ for which the best solution was obtained.}
  \label{tab:rez-kNN}
  \begin{tabularx}{\textwidth}{CCC}
  \toprule
    \textbf{Dataset} & \textbf{Best} & \boldmath{$k$}\\ \midrule
    ROST-P-1 &53.06\%& 8\\
    ROST-P-2 &54.08\%& 23\\
    ROST-P-3 &48.97\% & 11\\ \midrule
    ROST-PA-1 &31.63\%& 1\\
    ROST-PA-2 & 32.6\%& 1\\
    ROST-PA-3 &35.71\% & 1\\ \midrule
    ROST-PAC-1 &33.67\%& 2\\
    ROST-PAC-2 &\textbf{29.59\%} & 1\\
    ROST-PAC-3 &\textbf{29.59\%} & 4\\
  \bottomrule
  \end{tabularx}
\end{table}

\subsubsection{SVM}
\add[A.]{Prerequisite tests to determine the best kernel type and a good interval of values for the parameter $nu$ are described in Appendix~}\ref{app:prereq-results-SVM}, along with their results.

We ran tests for each kernel type and with \emph{nu} varying from 0.1 to 1, as we saw in Figure~\ref{fig:SVM-nu} that for values less than 0.1, SVM is unlikely to produce the best results. The best results obtained are shown in Table~\ref{tab:rez-SVM}. 

\begin{table}[H]
    \caption{SVM results 
    on the considered datasets. The number of incorrectly classified data is given as a percentage (the best results obtained by SVM on any ROST-*-* dataset representation are in bold)
    . \emph{Best} stands for the best test error rate (out of 30 runs with $nu$ ranging from 0.001 to 1), and \emph{nu} stands for the parameter specific to the selected type of SVM (i.e., nu-SVC). Results are given for each type of kernel that was used by the SVM. The best result obtained by SVM compared to all methods for a given ROST-X-n dataset representation is in a gray cell. 
    \label{tab:rez-SVM}}
	\begin{adjustwidth}{-\extralength}{0cm}
		\begin{tabularx}{\fulllength}{Ccccccccc}
			\toprule
& \multicolumn{2}{c}{\textbf{Linear Kernel}} & \multicolumn{2}{c}{\textbf{Polynomial Kernel}} & \multicolumn{2}{c}{\textbf{Radial Basis Kernel}}& \multicolumn{2}{c}{\textbf{Sigmoid Kernel}}\\
\textbf{Dataset} & \textbf{Best} &\emph{\textbf{nu}} & \textbf{Best} &\emph{\textbf{nu}} & \textbf{Best} &\emph{\textbf{nu}} & \textbf{Best} &\emph{\textbf{nu}} \\ \hline
\textbf{ROST-P-1} &
  \cellcolor{gray!25}43.87\% &$\ge$$ 0.6$ & 65.30\%&0.5 & 59.18\%&0.4 & 58.16\%&0.4 
\\
ROST-P-2 &
  55.10\%&$\ge$$0.6$ & 70.40\%&0.2, 0.4 & 67.34\%&0.4 & 68.37\%&0.2, 0.4
\\
ROST-P-3 &\cellcolor{gray!25}
  43.87\%&$\ge$$0.6$ & 65.30\%&0.5 & 59.18\%&0.4 & 58.16\%&0.4 
\\ \hline
ROST-PA-1 &31.63\%&0.5 & 51.02\%&0.5 & 44.89\%&0.3 & 45.91\%&0.3 
\\
ROST-PA-2 &26.53\%&0.5 & 55.10\%&$\ge$$0.6$ & 44.89\%&$\ge$$0.6$ & 44.89\%&$\ge$$0.6$ 
\\
ROST-PA-3 &
28.57\%&0.4 & 54.08\% &0.2, 0.3 & 51.02\%&0.2 & 51.02\%&0.2 
\\ \hline
\mbox{ROST-PAC-1 } 
  &\cellcolor{gray!25}\textbf{23.46}\%&0.2 & 54.08\%&0.2 & 50.00\%&0.5 & 50.00\%&0.5 
\\
\mbox{ROST-PAC-2 } & \cellcolor{gray!25}24.48\%&0.5 & 51.02\% &$\ge$$0.6$ & 39.79\% &$\ge$$0.6$ & 39.79\% &$\ge$$0.6$ 
\\
\mbox{ROST-PAC-3 } & 26.53\%&0.5 & 51.02\%&0.4 & 41.83\%&0.5 & 42.85\%&0.5 
\\
\bottomrule
		\end{tabularx}
	\end{adjustwidth}
\end{table}

As can be seen, the best values were obtained for values of parameter \emph{nu} between 0.2 and 0.6 (where sometimes 0.6 is the smallest value of the set \{0.6, 0.7, $\dots$, 1\} for which the best test error was obtained). 
The best value obtained by this method was $23.46\%$ for ROST-PAC-1, using the \emph{linear kernel} and \emph{nu} parameter value 0.2.

\subsubsection{Decision Trees with C5.0}
\add[A.]{Advanced pruning options for optimizing the decision trees with C5.0 model and their results are presented in Appendix~}\ref{app:prereq-results-DT}. \add[A.]{The best results were obtained by using $-m$ cases option, as detailed in Table~}\ref{tab:rez-DT-fin}.

\begin{table}[H]
\caption{Decision tree 
 results on the considered datasets. The number of incorrectly classified data is given as a percentage (the best results obtained by DT with C5.0 on any ROST-*-* dataset representation are in bold)
 . \emph{Error} stands for the test error rate, \textit{Size} stands for the size of the decision tree required for that specific solution and \emph{cases} stands for the threshold for which is decided to have two more that two branches at a specific branching point ($cases \in \{1, 2, \dots, 30\}$). The best result obtained by DT with C5.0 compared to all methods for a given ROST-X-n dataset representation is in a gray cell. 
 \label{tab:rez-DT-fin}}
		\begin{tabularx}{\textwidth}{C C C C}
		\toprule
\textbf{Dataset} & \textbf{Error}& \textbf{Size}&\textbf{Cases}\\ \midrule
\mbox{ROST-P-1} &51.0\%& 18 & 8\\
\mbox{ROST-P-2} &\cellcolor{gray!25}51.0\%& 46 & 3\\
\mbox{ROST-P-3} &57.1\%& 99 &1\\\midrule
\mbox{ROST-PA-1} &31.6\%& 13 &12\\
\mbox{ROST-PA-2} &26.5\%& 57 &1\\
\mbox{ROST-PA-3} &29.6\%& 31 &3\\\midrule
\mbox{ROST-PAC-1} &28.6\%& 39 &2\\
\mbox{ROST-PAC-2}&\cellcolor{gray!25} \textbf{24.5\%}& 12 &14\\
\mbox{ROST-PAC-3} &26.5\%& 13 &14\\
\bottomrule
		\end{tabularx}
\end{table}

The best result obtained by this method was $24.5\%$ on ROST-PAC-2, with $-m$ $14$ option, on a decision tree of size 12. When no options were used, the size of the decision trees was considerably larger for ROST-P-* (i.e., $\ge$57) than those for ROST-PA-* and ROST-PAC-* (i.e., $\le$39).


\subsection{Comparison and Discussion}
The findings of our investigations allow for a twofold perspective. The first perspective refers to the evaluation of the performance of the five investigated methods, as well as to the observation of the ability of the considered feature sets to better represent the dataset for successful classification. The other perspective is to place our results in the context of other state-of-the-art investigations in the field of author attribution.
\subsubsection{Comparing the Internally Investigated Methods}
From all the results presented above, upon consulting the tables containing the best test error rates, and especially the gray-colored cells (which contain the best results while comparing the methods amongst themselves) we can highlight the following:
\begin{itemize}
  \item \emph{ANN:} 
  \begin{itemize}
    \item Four best results for: ROST-PA-1, ROST-PA-3, ROST-PAC-2 and ROST-PAC-3 (see Table~\ref{tab:rez-ANN});
    \item Best ANN 23.46\% on ROST-PAC-3; best ANN average 36.93\% on ROST-PAC-2;
    \item Worst best \textbf{overall} $61.22\%$ on ROST-P-1.
  \end{itemize} 

  \item \emph{MEP:} 
  \begin{itemize}
    \item Two best results for ROST-PA-2 and ROST-PAC-3 (see Table~\ref{tab:rez-MEP});
    \item Best \textbf{overall} 20.40\% on ROST-PA-2; best \textbf{overall} average 27.95\% on ROST-PA-2;
    \item Worst best MEP $54.08\%$ on ROST-P-1.
  \end{itemize}

  \item \emph{k-NN:} 
  \begin{itemize}
    \item Zero best results (see Table~\ref{tab:rez-kNN});
    \item Best k-NN 29.59\% on ROST-PAC-2 and ROST-PAC-3; 
    \item Worst k-NN $54.08\%$ on ROST-P-2.
  \end{itemize}

  \item \emph{SVM:} 
  \begin{itemize}
    \item Four best results for: ROST-P-1, ROST-P-3, ROST-PAC-1 and ROST-PAC-2 (see \mbox{Table~\ref{tab:rez-SVM}});
    \item Best SVM 23.44\% on ROST-PAC-1; 
    \item Worst SVM $52.10\%$ on ROST-P-2.
  \end{itemize} 

  \item \emph{Decision trees:} 
  \begin{itemize}
    \item Two best results for: ROST-P-2 and ROST-PAC-2 (see Table~\ref{tab:rez-DT-fin});
    \item Best DT 24.5\% on ROST-PAC-2; 
    \item Worst DT $57.10\%$ on ROST-P-2.
  \end{itemize} 

\end{itemize}

Other notes from the results are:
\begin{itemize}
  \item Best values for each method were obtained for ROST-PA-2 or ROST-PAC-*;
  \item The worst of these best results were obtained for ROST-P-1 or ROST-P-2;
  \item ANN and MEP suffer from overfitting. The training errors are significantly smaller than the test errors. This problem can only be solved by adding more data to the training set.
\end{itemize}

An overview of the best test results obtained by all five methods is given in Table~\ref{tab:rez-discuss}. 
\begin{table}[H]
\caption{Top of methods 
 on each shuffle of each dataset, based on the best results achieved by each method. The gray-colored box represents the overall best (i.e., for all datasets and with all methods).
 \label{tab:rez-discuss}}
		\begin{tabularx}{\textwidth}{CCCCCC}
			\toprule
  \textbf{Dataset} & \textbf{1{st} Place} & \textbf{2{nd} Place}& \textbf{3{rd} Place}& \textbf{4{th} Place}& \textbf{5{th} Place}\\ \midrule
  ROST-P-1 & \textbf{SVM}& \textbf{DT}&\textbf{k-NN}& \textbf{MEP}& \textbf{ANN}\\
       & 43.87\% & 51.0\%&53.06\%& 54.08\%& 61.22\% \\
  ROST-P-2 & \textbf{DT}&\textbf{MEP}& \textbf{k-NN}& \textbf{SVM}& \textbf{ANN}\\
       & 51.0\%&52.04\%& 54.08\%& 55.10\%& 60.20\%\\
  ROST-P-3 &\textbf{SVM}&\multicolumn{2}{c}{\textbf{k-NN},\textbf{MEP}}& \multicolumn{2}{c}{\textbf{DT},\textbf{ANN}}\\
       & 43.87\%&\multicolumn{2}{c}{48.97\%}& \multicolumn{2}{c}{57.14\%}\\\midrule
  ROST-PA-1 & \textbf{ANN}&\textbf{MEP}& \multicolumn{3}{c}{\textbf{SVM},\textbf{DT},\textbf{k-NN}} \\
       & 24.48\% &29.59\%& \multicolumn{3}{c}{31.63\%}\\
  ROST-PA-2 &\cellcolor{gray!25}\textbf{MEP}& \textbf{ANN} &\multicolumn{2}{c}{\textbf{SVM},\textbf{DT}}& \textbf{k-NN}\\
       &\cellcolor{gray!25}20.40\%& 24.48\% & \multicolumn{2}{c}{26.53\%}& 32.6\%\\
  ROST-PA-3 &\textbf{ANN}&\textbf{SVM}&\multicolumn{2}{c}{\textbf{MEP},\textbf{DT}}& \textbf{k-NN}\\
        &26.53\%&28.57\%&\multicolumn{2}{c}{29.59\%} & 35.71\%\\\midrule
  ROST-PAC-1 & \textbf{SVM}&\textbf{ANN}& \textbf{MEP}& \textbf{DT}& \textbf{k-NN}\\
       & 23.46\%&24.48\%& 27.55\%& 28.6\%& 33.67\%\\
  ROST-PAC-2 & \multicolumn{3}{c}{\textbf{SVM},\textbf{DT},\textbf{ANN}} &\textbf{MEP} & \textbf{k-NN}\\
       &\multicolumn{3}{c}{ 24.48\%}&26.53\%& 29.59\%\\
  ROST-PAC-3 &\multicolumn{2}{c}{\textbf{MEP},\textbf{ANN}}& \multicolumn{2}{c}{\textbf{SVM},\textbf{DT}}& \textbf{k-NN}\\
       &\multicolumn{2}{c}{23.46\%}&\multicolumn{2}{c}{26.53\%}& 29.59\%\\
\bottomrule
		\end{tabularx}
\end{table}

ANN ranks last for all ROST-P-* and ranks 1st and 2nd for ROST-PA-* and ROST-PAC-*. MEP is either ranked 1st or ranked 2nd on all ROST-*-* with three exceptions, i.e., for ROST-P-1 and ROST-PAC-2 (at 4th place) and for ROST-PAC-1 (at 3rd place). k-NN performs better (i.e., 3rd and 2nd places) on ROST-P-*, and ranks last for ROST-PA-* and ROST-PAC-*. SVM is ranked 1st for ROST-P-* and ROST-PAC-* with two exceptions: for ROST-P-2 (ranked 4th) and for ROST-PAC-3 (on 3rd place). For ROST-PA-* SVM is in 3rd and 2nd places. Decision trees (DT) with C5.0 is mainly on the 3rd and 4th places, with three exceptions: for ROST-P-1 (on 2nd place), for ROST-P-2 (on 1st place), and for ROST-PAC-2 (on 1st place). 

An overview of the average test results obtained by all five methods is given in Table~\ref{tab:rez-discussAvg}. However, for ANN and MEP alone, we could generate different results with the same parameters, based on different starting \emph{seed} values, with which we ran 30 different runs. For the other 3 methods, we used the best results obtained with a specific set of parameters (as in Table~\ref{tab:rez-discuss}).

Comparing all 5 methods based on averages, SVM and DT take the lead as the two methods that share the 1st and 2nd places with two exceptions, i.e., for ROST-P-2 and ROST-P3 for which SVM and DT, respectively, rank 3rd. k-NN usually ranks 3rd, with four exceptions, when k-NN was ranked 2nd for ROST-P-2 and ROST-P-3, for ROST-PA-1 for which k-NN ranks 1st together with SVM and DT, and for ROST-PA-2 for which k-NN ranks 4th. MEP is generally ranked 4th with one exception, i.e., for ROST-PA-2 for which it ranks 3rd. ANN ranks last for all ROST-*-*.

\begin{figure}[H]
  
  \hspace{-8pt}\begin{subfigure}[b]{\textwidth}
     \centering
     \includegraphics[width=\textwidth]{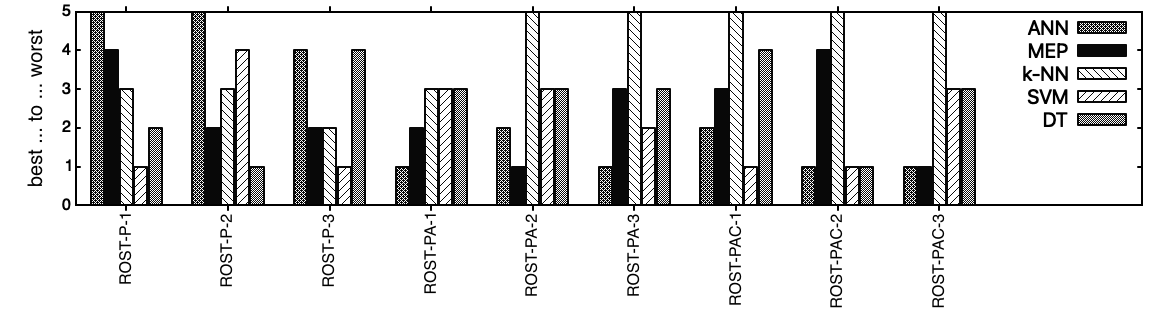}
     \caption{\centering}
     \label{fig:rez-disc}
   \end{subfigure}\\

    \hspace{-8pt}\begin{subfigure}[b]{\textwidth}
     \centering
     \includegraphics[width=\textwidth]{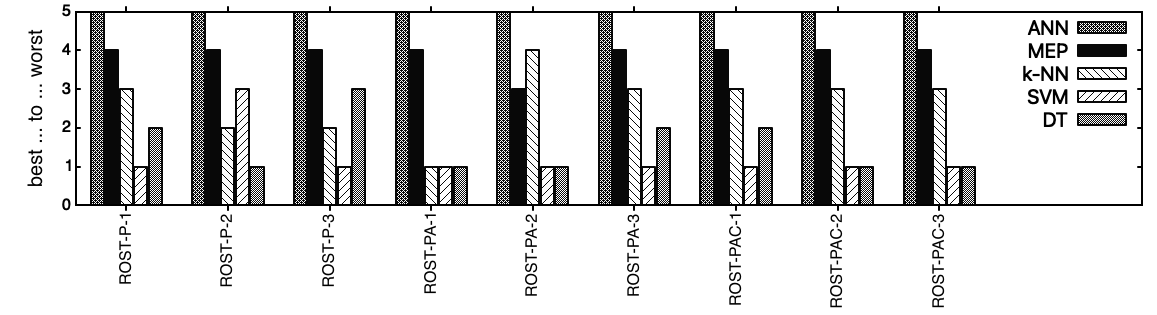}
     \caption{\centering}
     \label{fig:rez-disc-avg}
   \end{subfigure}
  \caption{Top of methods on each shuffle of each dataset. Lower values are better. (\textbf{a}) Top of best results obtained by all methods (\textbf{b}) Top of average results, when applicable (i.e., over 30 runs for ANN and MEP).}
  \label{fig:rez-discuss}
\end{figure}

For a better visual representation, we have plotted the results from Tables~\ref{tab:rez-discuss} and~\ref{tab:rez-discussAvg} in Figure~\ref{fig:rez-discuss}.
\begin{table}[H]
\caption{Top of methods 
 on average results on each shuffle of each dataset. For k-NN, SVM, and DT we do not have 30 runs with the same parameters, so for these methods, the best values are presented here. The gray-colored box represents the overall best average (i.e., on all datasets and with all methods).\label{tab:rez-discussAvg}}
		\begin{tabularx}{\textwidth}{CCCCCC}
					\toprule
  \textbf{Dataset} & \textbf{1{st} Place} & \textbf{2{nd} Place}& \textbf{3{rd} Place}& \textbf{4{th} Place}& \textbf{5{th} Place}\\ \midrule
  ROST-P-1 & \textbf{SVM}& \textbf{DT}&\textbf{k-NN}& \textbf{MEP}& \textbf{ANN}\\
       & 43.87\% & 51.0\%&53.06\%& 61.32\%& 76.70\% \\
  ROST-P-2 & \textbf{DT}& \textbf{k-NN}& \textbf{SVM}&\textbf{MEP}& \textbf{ANN}\\
       & 51.0\%& 54.08\%& 55.10\%&62.51\%& 80.27\%\\
  ROST-P-3 &\textbf{SVM}&\textbf{k-NN}& \textbf{DT} &\textbf{MEP}& \textbf{ANN}\\
       & 43.87\%&48.97\%& 57.14\%& 58.84\%& 80.95\%\\\midrule
  ROST-PA-1 & \multicolumn{3}{c}{\textbf{SVM},\textbf{DT},\textbf{k-NN}} &\textbf{MEP}& \textbf{ANN}\\
        & \multicolumn{3}{c}{31.63\%}&36.49\%& 45.03\%\\
  ROST-PA-2 &\multicolumn{2}{c}{\textbf{SVM},\textbf{DT}}&\textbf{MEP} & \textbf{k-NN}& \textbf{ANN}\\
       & \multicolumn{2}{c}{26.53\%}& 27.95\%&32.6\%& 41.73\% \\
  ROST-PA-3 &\textbf{SVM}&\textbf{DT}& \textbf{k-NN}&\textbf{MEP}&\textbf{ANN}\\
        &28.57\%&29.59\% & 35.71\%&39.93\%&47.82\%\\\hline
  ROST-PAC-1 & \cellcolor{gray!25}\textbf{SVM}& \textbf{DT}& \textbf{k-NN}& \textbf{MEP}&\textbf{ANN}\\
       & \cellcolor{gray!25}23.46\%& 28.6\%& 33.67\%& 33.84\%&38.16\%\\
  ROST-PAC-2 & \multicolumn{2}{c}{\textbf{SVM},\textbf{DT}} & \textbf{k-NN}&\textbf{MEP} &\textbf{ANN}\\
       &\multicolumn{2}{c}{ 24.48\%}& 29.59\%&34.89\%& 36.93\%\\
  ROST-PAC-3& \multicolumn{2}{c}{\textbf{SVM},\textbf{DT}}& \textbf{k-NN} &\textbf{MEP}&\textbf{ANN}\\
       &\multicolumn{2}{c}{26.53\%}& 29.59\%&34.38& 37.21\%\\
\bottomrule
		\end{tabularx}
\end{table}
\unskip

We performed statistical tests to determine whether the results obtained by MEP and ANN are significantly different with a 95\% confidence level. The tests were two-sample, equal variance, and two-tailed T-tests. The results are shown in Table~\ref{tab:my_pValue}. 
\begin{table}[H] 
\caption{\emph{p}-values obtained when comparing MEP and ANN results over 30 runs. \emph{No. of neurons used by ANN on the hidden layer} represents the best-performing ANN structure on the specific ROST-*-*.\label{tab:my_pValue}}
\begin{tabularx}{\textwidth}{CCC}
\toprule
\textbf{Dataset} &\textbf{\emph{p}-Value (ANN vs. MEP Results)} & \textbf{No. of Neurons Used by ANN on the Hidden Layer}\\ \midrule
ROST-P-1 & 1.98 $\times~10^{-15}$ & 46 \\ 
ROST-P-2 & 4.23 $\times~10^{-11}$ & 36 \\ 
ROST-P-3 & 3.86 $\times~10^{-15}$ & 28 \\ \midrule
ROST-PA-1 & 1.14 $\times~10^{-5}$ & 40 \\ 
ROST-PA-2 & 6.57 $\times~10^{-15}$ & 45 \\ 
ROST-PA-3 & 3.07 $\times~10^{-4}$ & 27 \\ \midrule
ROST-PAC-1 & 2.47 $\times~10^{-4}$ & 49 \\ 
ROST-PAC-2 & 1.07 $\times~10^{-1}$ & 40 \\ 
ROST-PAC-3 & 2.80 $\times~10^{-2}$ & 41\\ 
\bottomrule
\end{tabularx}
\end{table}

The \emph{p}-values obtained show that the MEP and ANN test results are statistically significantly different for almost all ROST-*-* (i.e., $p<0.05$) with one exception, i.e., for ROST-PAC-2 for which the differences are not statistically significant (i.e., $p=0.107$).

\begin{figure}[H]
   \hspace{-8pt}
\begin{subfigure}[b]{\textwidth}
     \centering
     \includegraphics[width=0.69\textwidth]{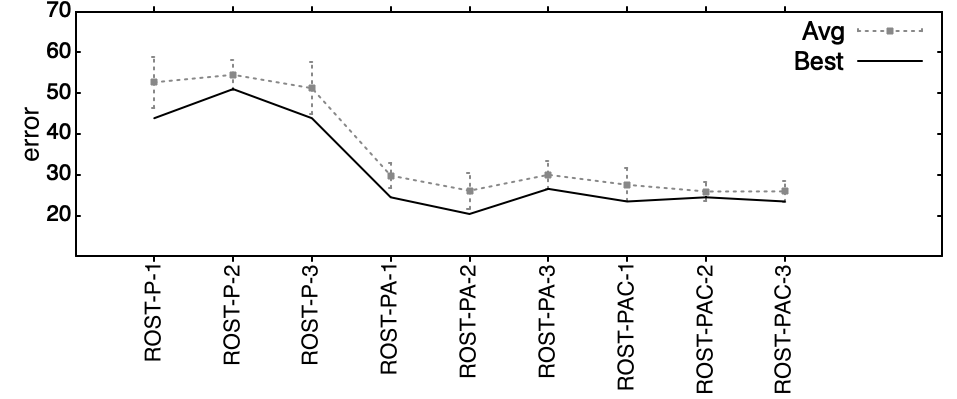}
     \includegraphics[width=0.29\textwidth]{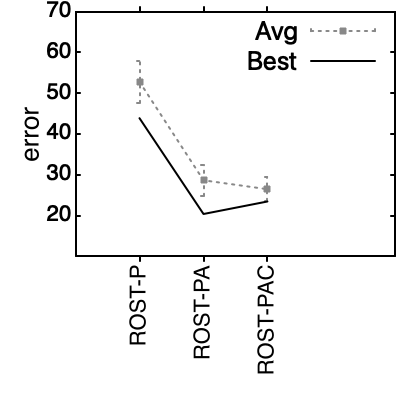}
     \caption{ \centering}
     \label{fig:rez-o}
   \end{subfigure}\\
 
   \hspace{-8pt}\begin{subfigure}[b]{\textwidth}
     \centering
     \includegraphics[width=0.69\textwidth]{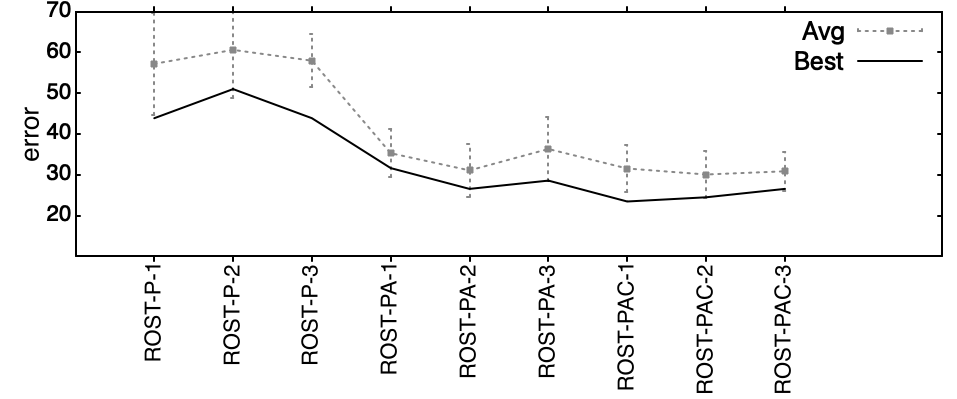}
     \includegraphics[width=0.29\textwidth]{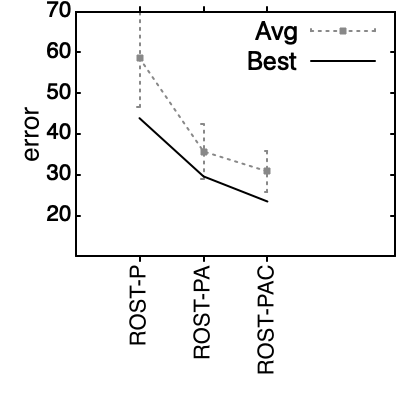}
     \caption{ \centering}
     \label{fig:rez-o-avg}
   \end{subfigure}
    \caption{Results on the best solutions obtained on the considered datasets. The percentage of incorrectly classified data is plotted. \textit{Best} stands for the best solution, \textit{Avg} stands for \textit{Average} and the \textit{Standard Deviation} is represented by error bars. (\textbf{a}) \textit{Best}, \textit{Average} and \textit{Standard Deviation} are computed on the values from Table~\ref{tab:rez-discuss}; (\textbf{b}) \textit{Best}, \textit{Average}, and \textit{Standard Deviation} are computed on the values given in Table~\ref{tab:rez-discussAvg}.} 
    \label{fig:rez-discOverall}
\end{figure}

Next, we wanted to see which feature set, out of the three we used, was the best for successful author attribution. Therefore, we plotted all best and best average results obtained with all methods (as presented in Tables~\ref{tab:rez-discuss} and~\ref{tab:rez-discussAvg}) on all ROST-*-* and aggregated on the three datasets corresponding to the distinct feature lists, in Figure~\ref{fig:rez-discOverall}.

Based on the results represented in Figure~\ref{fig:rez-discOverall}a (i.e., which considered only the best results, as detailed in Table~\ref{tab:rez-discuss}) we can conclude that we obtained the best results on ROST-PA-* (i.e., corresponding to the 415 feature set, which contains prepositions and adverbs). However, using the average results, as shown in Figure~\ref{fig:rez-discOverall}b and detailed in Table~\ref{tab:rez-discussAvg} we infer that the best performance is obtained on ROST-PAC-* (i.e., corresponding to the 432-feature set, containing prepositions, adverbs, and conjunctions).

Another aspect worth mentioning based on the graphs presented in Figure~\ref{fig:rez-discOverall} is related to the standard deviation (represented as error bars) between the results obtained by all methods considered on all considered datasets. Standard deviations are the smallest in Figure~\ref{fig:rez-discOverall}a, especially for ROST-PA-* and even more so for ROST-PAC-*. This means that the methods perform similarly on those datasets. For ROST-P-* and in Figure~\ref{fig:rez-discOverall}b, the standard deviations are larger, which means that there are bigger differences between the methods.

\subsubsection{Comparisons with Solutions Presented in Related Work}
To better evaluate our results and to better understand the discriminating power of the best performing method (i.e., MEP on ROST-PA-2), we also calculate the \emph{macro-accuracy} (or \emph{macro-average accuracy}). This metric allows us to compare our results with the results obtained by other methods on other datasets, as detailed in Table~\ref{tab:StateOfArt}. For this, we considered the test for which we obtained our best result with MEP, 
with a test error rate of 20.40\%. \add[A.]{This means that 20 out of 98 tests were misclassified.}

To perform all the necessary calculations, two vectors are required, i.e., the vector of \emph{targets} or \emph{true values} (i.e., authors/classes that were the actual authors 
of the test texts) and the vector of \emph{outputs} or \emph{predicted values} (i.e., authors/classes identified by the algorithm as the authors of the test texts). 
The resulting \emph{Confusion Matrix} is depicted in Table~\ref{tab:confusion_matrix}.

\begin{table}[H]
    \caption{\emph{Confusion Matrix} 
   (on the right side). 
   Column headers and row headers (i.e., numbers from 0 to 9 that are written in bold) 
   are the codes\textsuperscript{1} given to our authors, as specified \change[A.]{in the first columns of Tables~6--8}{on the left side}.}
  \label{tab:confusion_matrix}

  \begin{tabularx}{\textwidth}{c l C c|c c c c c c c c c c}
  \toprule
\textbf{Code} & \textbf{Author} &&& \textbf{0} & \textbf{1} & \textbf{2} & \textbf{3} & \textbf{4} & \textbf{5} & \textbf{6} & \textbf{7} & \textbf{8} & \textbf{9} \\\midrule 
0 & Ion \textbf{Creangă} &&\textbf{0} & 6 & 0 & 0 & 0 & 0 & 0 & 0 & 0 & 0 & 1 \\ 
1& Barbu Ştefănescu \textbf{Delavrancea} &&\textbf{1} & 0 & 4 & 0 & 3 & 1 & 0 & 1 & 0 & 0 & 2 \\ 
2&Mihai \textbf{Eminescu} && \textbf{2} & 0 & 0 & 6 & 0 & 0 & 0 & 0 & 0 & 0 & 0 \\ 
3&Nicolae \textbf{Filimon} && \textbf{3} & 0 & 1 & 1 & 6 & 0 & 0 & 0 & 0 & 0 & 0 \\ 
4&Emil \textbf{Gârleanu} && \textbf{4} & 1 & 1 & 0 & 0 & 6 & 0 & 0 & 0 & 1 & 1 \\ 
5 &Petre \textbf{Ispirescu} && \textbf{5} & 0 & 0 & 0 & 0 & 0 & 10 & 0 & 0 & 0 & 0 \\ 
6& Mihai \textbf{Oltean} && \textbf{6} & 0 & 0 & 0 & 0 & 0 & 0 & 8 & 0 & 0 & 0 \\ 
7&Emilia \textbf{Plugaru} && \textbf{7} & 0 & 1 & 0 & 0 & 1 & 0 & 0 & 8 & 0 & 0 \\ 
8&Liviu \textbf{Rebreanu} && \textbf{8} & 0 & 1 & 0 & 0 & 0 & 0 & 1 & 0 & 12 & 1 \\ 
9&Ioan \textbf{Slavici} && \textbf{9} &0 & 0 & 0 & 0 & 0 & 0 & 0 & 0 & 1 & 12 \\ 
\bottomrule
  \end{tabularx}
  	\noindent{\footnotesize{\textsuperscript{1} 
  	The authors' codes are the same as those specified in the first columns of Tables~\ref{tab:textStats2}
--\ref{tab:textDiv}.}}

\end{table}

This matrix is a representation that highlights for each class/author the \emph{true positives} (i.e., the number of cases in which an author was correctly identified as the author of the text), the \emph{true negatives} (i.e., the number of cases where an author was correctly identified as not being the author of the text), the \emph{false positives} (i.e., the number of cases in which an author was incorrectly identified as being the author of the text), the \emph{false negatives} (i.e., the number of cases where an author was incorrectly identified as not being the author of the text). For binary classification, these four categories are easy to identify. However, in a multiclass classification, the \emph{true positives} are contained in the main diagonal cells corresponding to each author, but the other three categories are distributed according to the actual authorship attribution made by the algorithm. 

For each class/author, various metrics are calculated based on the confusion matrix. They are:
\begin{itemize}
  \item \emph{Precision}---the number of correctly attributed authors divided by the number of instances when the algorithm identified the attribution as correct;
  \item \emph{Recall (Sensitivity)}---the number of correctly attributed authors divided by the number of test texts belonging to that author;
  \item \emph{F-score}---a combination of the \emph{Precision} and \emph{Recall (Sensitivity)}.
\end{itemize}

Based on these individual values, classification performance metrics 
are calculated. The overall results are shown in Table~\ref{tab:AccuracyEvaluationResults}.
\begin{table}[H] 
\caption{Accuracy evaluation 
 Results.  
 \label{tab:AccuracyEvaluationResults}}
 \begin{tabularx}{\textwidth}{r C C C l}
     \toprule
     class&\textbf{precision}  &  \textbf{recall} & \textbf{f1-score} &  support \\\midrule 

        0 & 0.857 & 0.857 & 0.857 &    7\\
        1 & 0.500 & 0.364 & 0.421 &   11\\
        2 & 0.857 & 1.000 & 0.923 &    6\\
        3 & 0.667 & 0.750 & 0.706 &    8\\
        4 & 0.750 & 0.600 & 0.667 &   10\\
        5 & 1.000 & 1.000 & 1.000 &   10\\
        6 & 0.800 & 1.000 & 0.889 &    8\\
        7 & 1.000 & 0.800 & 0.889 &   10\\
        8 & 0.857 & 0.800 & 0.828 &   15\\
        9 & 0.706 & 0.923 & 0.800 &   13\\\midrule 
        
        \textbf{accuracy} &  &   &  0.796 &    98\\
        \textbf{macro avg} &  0.799 & 0.809 & 0.798 &    98\\
        \textbf{weighted avg} &  0.795 & 0.796 & 0.789 &    98\\
    \bottomrule
 \end{tabularx}
\end{table}
These results are obtained by using sklearn.metrics's \emph{classification\_report}~\cite{classificationReport}. In order to obtain the \emph{macro-accuracy}, sklearn.metrics's \emph{balanced\_accuracy\_score}~\cite{balancedAccuracyScore} was used, obtaining a value of 80.94\%.

We have 400 documents, and 10 authors in our dataset. The \emph{content} of our texts is 
\emph{cross-genre} (i.e., stories, short stories, fairy tales, novels, articles, and sketches) and \emph{cross-topic} (as in different texts, different topics are covered). We also calculated an average number of words per document, which is 3355, and the \emph{imbalance} (considered in~\cite{tyo2022state} to be the standard deviation of the number of documents per author), which in our case is 10.45. Our type of investigation can be considered to be part of the Ngram class (this class and other investigation-type classes are presented in Section~\ref{related_work-comparison}). Next, we recreated Table~\ref{tab:StateOfArt} (depicted in Section~\ref{related_work-comparison}) while reordering the datasets based on their macro-accuracy results obtained by Ngram class methods in reverse order, and we have appropriately placed details of our own dataset and the macro-accuracy we achieved with MEP as shown above. This top is depicted in Table~\ref{tab:StateOfArtAndROST}.

\begin{table}[H]
\caption{State of the art 
 \emph{macro-accuracy} of authorship attribution models. Information collected from~\cite{tyo2022state} (Tables \ref{tab:datasets} and \ref{tab:datasetDivSize}). 
  \emph{Name} is the name of the dataset; \emph{No.docs} represents the number of documents in that dataset; \emph{No. auth} represents the number of authors; \emph{Content} indicates whether the documents are cross-topic ($\times_t$) or cross-genre ($\times_g$); \emph{W/D} stands for \emph{words per documents}, being the average length of documents; \emph{imb} represents the \emph{imbalance} of the dataset as measured by the standard deviation of the number of documents per author.
  \label{tab:StateOfArtAndROST}}
	\begin{adjustwidth}{-\extralength}{0cm}
\setlength{\cellWidtha}{\fulllength/10-2\tabcolsep+0.2in}
\setlength{\cellWidthb}{\fulllength/10-2\tabcolsep-0in}
\setlength{\cellWidthc}{\fulllength/10-2\tabcolsep-0in}
\setlength{\cellWidthd}{\fulllength/10-2\tabcolsep-0in}
\setlength{\cellWidthe}{\fulllength/10-2\tabcolsep-0in}
\setlength{\cellWidthf}{\fulllength/10-2\tabcolsep-0in}
\setlength{\cellWidthg}{\fulllength/10-2\tabcolsep-0in}
\setlength{\cellWidthh}{\fulllength/10-2\tabcolsep-0.2in}
\setlength{\cellWidthi}{\fulllength/10-2\tabcolsep-0in}
\setlength{\cellWidthj}{\fulllength/10-2\tabcolsep-0in}
\scalebox{1}[1]{\begin{tabularx}{\fulllength}{>{\centering\arraybackslash}m{\cellWidtha}>{\centering\arraybackslash}m{\cellWidthb}>{\centering\arraybackslash}m{\cellWidthc}>{\centering\arraybackslash}m{\cellWidthd}>{\centering\arraybackslash}m{\cellWidthe}>{\centering\arraybackslash}m{\cellWidthf}>{\centering\arraybackslash}m{\cellWidthg}>{\centering\arraybackslash}m{\cellWidthh}>{\centering\arraybackslash}m{\cellWidthi}>{\centering\arraybackslash}m{\cellWidthj}}
			\toprule
    \multicolumn{6}{c}{\textbf{Dataset }} & \multicolumn{4}{c}{\textbf{Investigation Type} }\\\midrule
    \textbf{Name} & \textbf{No. Docs} & \textbf{No. Auth} & \textbf{Content}& \textbf{W/D}& \textbf{Imb}& \textbf{Ngram} & \textbf{PPM} & \textbf{BERT} & \textbf{pALM}\\\midrule
    
    \textcolor{gray!75}{Guardian10}&\textcolor{gray!75}{ 444}&\textcolor{gray!75}{13 }&\textcolor{gray!75}{ $\times_t$ $\times_g$}&\textcolor{gray!75}{ 1052}&\textcolor{gray!75}{ 6.7}&\textcolor{gray!75}{ 100 }&\textcolor{gray!75}{ 86.28 }&\textcolor{gray!75}{ 84.23 }&\textcolor{gray!75}{ 66.67 }\\
    
    \textcolor{gray!75}{IMDb62 }&\textcolor{gray!75}{ 62,000}&\textcolor{gray!75}{62 }&\textcolor{gray!75}{ $-$}&\textcolor{gray!75}{ 349}&\textcolor{gray!75}{ 2.6}&\textcolor{gray!75}{ 98.81 }&\textcolor{gray!75}{ 95.90 }&\textcolor{gray!75}{ 98.80 }&\textcolor{gray!75}{ $-$}\\ 
    
    
    \textcolor{gray!75}{CMCC }&\textcolor{gray!75}{ 756}&\textcolor{gray!75}{21}&\textcolor{gray!75}{ $\times_t$ $\times_g$}&\textcolor{gray!75}{ 601}&\textcolor{gray!75}{ 0}&\textcolor{gray!75}{ 86.51 }&\textcolor{gray!75}{ 62.30 }&\textcolor{gray!75}{ 60.32 }&\textcolor{gray!75}{ 54.76}\\\midrule
    
    ROST & 400&10& $\times_t$ $\times_g$&3355& 10.45 &80.94& $-$ & $-$ & $-$ \\\midrule
    
    \textcolor{gray!75}{CCAT50} &\textcolor{gray!75}{ 5000}&\textcolor{gray!75}{50}&\textcolor{gray!75}{ $-$ }&\textcolor{gray!75}{ 506}&\textcolor{gray!75}{ 0}&\textcolor{gray!75}{ 76.68}&\textcolor{gray!75}{ 69.36 }&\textcolor{gray!75}{ 65.72 }&\textcolor{gray!75}{ 63.36}\\
    
    \textcolor{gray!75}{Blogs50 }&\textcolor{gray!75}{ 66,000}&\textcolor{gray!75}{50}&\textcolor{gray!75}{$-$ }&\textcolor{gray!75}{ 122}&\textcolor{gray!75}{ 553}&\textcolor{gray!75}{ 72.28}&\textcolor{gray!75}{ 72.16 }&\textcolor{gray!75}{ 74.95 }&\textcolor{gray!75}{ $-$ }\\
    
    \textcolor{gray!75}{PAN20 }&\textcolor{gray!75}{ 443,000}&\textcolor{gray!75}{278,000 }&\textcolor{gray!75}{$\times_t$ }&\textcolor{gray!75}{ 3922}&\textcolor{gray!75}{2.3}&\textcolor{gray!75}{ 43.52 }&\textcolor{gray!75}{ $-$ }&\textcolor{gray!75}{ 23.83 }&\textcolor{gray!75}{ $-$}\\
    
    \textcolor{gray!75}{Gutenburg }&\textcolor{gray!75}{ 28,000}&\textcolor{gray!75}{4500}&\textcolor{gray!75}{ $-$ }&\textcolor{gray!75}{ 66,350}&\textcolor{gray!75}{ 10.5}&\textcolor{gray!75}{57.69}&\textcolor{gray!75}{ $-$ }&\textcolor{gray!75}{ 59.11 }&\textcolor{gray!75}{ $-$ }\\
    \bottomrule
		\end{tabularx}}
	\end{adjustwidth}
\end{table}

\remove[A.]{Based on these results, the MEP method ranks 3$^rd$ among the other methods presented in related work. However, w}We would like to underline the large imbalance of our dataset compared with the first 
three datasets, the fact that we had fewer documents, and the fact that the average number of words in our texts, although higher, has a large standard deviation, as already shown in \change[A.]{Figure~2}{Table}~\ref{tab:datasetDivSize}
. Furthermore, as already presented in Section~\ref{proposed_dataset}, our dataset is by design very heterogeneous from multiple perspectives which are not only in terms of content and size, but also the differences that pertain to the time periods of authors, the medium they wrote for (paper or online media), and the sources of the texts. Although all these aspects do not restrict the new test texts to certain characteristics (to be easily classified by the trained model), they make the classification problem even harder.

\section{Conclusions and Further Work}
\label{future_work}

In this paper, we introduced a new dataset of Romanian texts \remove[E.E.]{written} by different authors. This dataset is heterogeneous from multiple perspectives, \change[E.E.]{, i.e., }{such as} the length of the texts, the sources from which they were collected, the time period in which the authors lived and wrote these texts, the \change[E.E.]{medium intended to be read from}{intended reading medium} (i.e., paper or online), and the type of writing (i.e., stories, short stories, fairy tales, novels, literary articles, and sketches). By choosing these very diverse texts we wanted to \change[E.E.]{have a more naturally selected collection that would not restrict new texts to any of these coordinates}{make sure that the new texts do not have to be restricted by these constraints}. As features, we wanted to use the \emph{inflexible parts of speech} (i.e., those that do not change their form in the context of communication): conjunctions, prepositions, interjections, and adverbs. After a closer investigation of their relevance to our dataset, we decided to use only prepositions, adverbs, and conjunctions, in that specific order, thus having three different feature lists of (1) 56 prepositions; (2) 415 prepositions and adverbs; and (3) 432 prepositions, adverbs, and conjunctions. Using these features, we constructed a numerical representation of our texts as vectors containing the frequencies of occurrence of the features in the considered texts, thus obtaining 3 distinct representations of our initial dataset. We divided the texts into training--validation--test sets of 50\%--25\%--25\% ratios, while randomly shuffling them three times in order to have three randomly selected arrangements of texts in each set of training, validation, and testing.  

To build our classifiers, we used five artificial intelligence techniques, namely artificial neural networks (ANN), multi-expression programming (MEP), k-nearest neighbor (k-NN), support vector machine (SVM), and decision trees (DT) with C5.0. We used the trained classifiers for authorship attribution on the texts selected for the test set. The best result we obtained was with MEP. By using this method, we obtained an overall ``best'' on all shuffles and all methods, which is of a $20.40\%$ error rate. 

Based on the results, we tried to determine which of the three distinct feature lists lead to the best performance. This inquiry was twofold. First, we considered the \emph{best} results obtained by all methods. From this perspective, we achieved the best performance when using ROST-PA-* (i.e., the dataset with 415 features, which contains prepositions and adverbs). Second, we considered the \emph{average} results over 30 different runs for ANN and MEP. These results indicate that the best performance was achieved when using ROST-PAC-* (i.e., the dataset with 432 features, which contains prepositions, adverbs, and conjunctions).

We also calculated the macro-accuracy for the best MEP result to compare it with other state-of-the-art methods on other datasets. 

\add[A.]{Given all the trained models that we obtained, the first future work is using ensemble decision. Additionally, determining whether multiple classifiers made the same error (i.e., attributing one text to the same incorrect author instead of the correct one) may mean that two authors have a similar style. This investigation can also go in the direction of detecting style similarities or grouping authors into style classes based on such similarities. }

\remove[A.]{We would like to continue our investigation using simpler feature sets that do not involve Natural Language processing. This decision is twofold. Firstly because, as we presented in the related work section, simpler methods usually give better results, while (secondly) more complex feature selection approaches (i.e., syntactic and semantic) are not that accurate because they usually introduce noise due to errors generated by complex text processing. }

\change[A.]{Furthermore, we would like to diversify our area of investigation, not only by fine-tuning}{Extending our area of research is also how we would like to continue our investigations.
We will not only fine-tune} the current methods but also expand to the use of recurrent neural networks (RNN) and convolutional neural networks (CNN). 

Regarding fine-tuning, we have already started an investigation using the top $N$ most frequently used words in our corpus. Even though we have some preliminary results, this investigation is still a work in progress. 

\change[E.E.]{Deep learning is also a direction we would like to investigate as a fine-tuning of ANN.}{Using deep learning to fine-tune ANN is another direction we would like to tackle.} We would also like to address overfitting and find solutions to mitigate this problem. 

\add[A.]{Linguistic analysis could help us as a complementary tool for detecting peculiarities that pertain to a specific author. For that, we will consider using long short-term memory (LSTM) architectures and pre-trained BERT models that are already available for Romanian. However, considering that a large section of our texts was written one or two centuries ago, we might need to further train BERT to be able to use it in our texts. That was one reason that we used inflexible parts of speech, as the impact of the diachronic developments of the language was greatly reduced. }

We would also \change[E.E.]{try to see the results that we obtain by considering}{investigate} the profile-based approach, where texts are treated cumulatively (per author) to build a \emph{profile}, which is a representation of the author’s style. Up to this point we have treated the training texts individually, an approach called \emph{instance-based}.

In terms of moving towards other types of neural networks, we would like to achieve the initial idea from which this entire area of research was born, namely finding a ``fingerprint'' of an author. We already have some incipient ideas on how these instruments may help us in our endeavor, but these new directions are still in the very early stages for us.

Improving upon the dataset is also high on our priority list. We are considering adding new texts and new authors. 
\vspace{6pt}

\authorcontributions{Conceptualization, S.M.A.; methodology, S.M.A. and M.O.; software, S.M.A. and M.O.; validation, S.M.A. and M.O.; formal analysis, S.M.A.; investigation, S.M.A.; resources, S.M.A.; data curation, S.M.A.; writing---original draft preparation, S.M.A.; writing---review and editing, S.M.A. and M.O.; visualization, S.M.A.; supervision, M.O.; project administration, S.M.A.; funding acquisition, S.M.A. All authors have read and agreed to the published version of the manuscript.}

\funding{
This research received no external funding. 
}

\institutionalreview{Not applicable.}

\informedconsent{Not applicable.}

\dataavailability{The proposed, used and analyzed dataset is available at 
 \url{https://www.kaggle.com/datasets/sandamariaavram/rost-romanian-stories-and-other-texts}. The source code that we wrote to perform the tests are available at \href{https://github.com/sanda-avram/ROST-source-code}{https://github.com/sanda-avram/ROST-source-code}, accessed on 28 November 2022. The data presented in Tables~\ref{tab:StateOfArt} and~\ref{tab:StateOfArtAndROST} are openly available in [\href{https://arxiv.org/abs/2209.06869v2}{arXiv:2209.06869v2}] at  \href{https://doi.org/10.48550/arXiv.2209.06869}{https://doi.org/10.48550/arXiv.2209.06869}, accessed on 29 November 2022.} 

\acknowledgments{We thank Ludmila Jahn who helped with the English revision of the text.
\\
\\
For calculating the \emph{macro-accuracy}, we initially used the \emph{Accuracy evaluation} tool available at~\cite{accuracyEvaluation}, build based on the paper~\cite{sokolova2009systematic}, obtaining a value of 88.84\%. Continuing our work we come to the realisation that this value was in fact erroneous. We, thereafter used a number of other methods to compute \emph{macro-accuracy}, which all gave us the value of 80.94\%. In this version of the paper we present the python module solution by using sklearn.metrics's \emph{classification\_report}~\cite{classificationReport} and  \emph{balanced\_accuracy\_score}~\cite{balancedAccuracyScore} .
}

\conflictsofinterest{
The authors declare no conflict of interest.
} 

\abbreviations{Abbreviations}{
The following abbreviations are used in this manuscript:\\

\noindent 
\begin{tabular}{@{}ll}
AA & Authorship Attribution\\
ANN & Artificial Neural Networks\\
MPE & Multi Expression Programming\\
k-NN & k-Nearest Neighbour\\
SVM & Support Vector Machines\\
DT & Decision Trees\\
RMSE & Root Mean Square Error\\
FANN & Fast Artificial Neural Network \\
MEPX & Multi Expression Programming software\\
LIBSVM & Support Vector Machine library \\
C5.0 & system for classifiers in the form of decision trees and rulesets\\
PoS & Part of Speech\\ 
IPoS & Inflexible Part of Speech\\ 
ROST & ROmanian Stories and other Texts\\
ROST-P-1 & ROST dataset using prepositions as features, shuffle 1\\
ROST-P-2~~~~~~ & ROST dataset using prepositions as features, shuffle 2\\
ROST-P-3 & ROST dataset using prepositions as features, shuffle 3\\
\end{tabular}

\noindent 
\begin{tabular}{@{}ll}

ROST-P-* & ROST-P-1 and ROST-P-2 and ROST-P-3\\
ROST-PA-1 & ROST dataset using prepositions and adverbs as features, shuffle 1\\
ROST-PA-2 & ROST dataset using prepositions and adverbs as features, shuffle 2\\
ROST-PA-3 & ROST dataset using prepositions and adverbs as features, shuffle 3\\
ROST-PA-* & ROST-PA-1 and ROST-PA-2 and ROST-PA-3\\
ROST-PAC-1 & ROST dataset using prepositions, adverbs, and conjunctions as features, shuffle 1\\
ROST-PAC-2 & ROST dataset using prepositions, adverbs, and conjunctions as features, shuffle 2\\
ROST-PAC-3 & ROST dataset using prepositions, adverbs, and conjunctions as features, shuffle 3\\
ROST-PAC-* & ROST-PAC-1 and ROST-PAC-2 and ROST-PAC-3\\
ROST-PC-1 & ROST dataset using prepositions and conjunctions as features, shuffle 1\\
ROST-PC-2 & ROST dataset using prepositions and conjunctions as features, shuffle 2\\
ROST-PC-3 & ROST dataset using prepositions and conjunctions as features, shuffle 3\\
ROST-*-* & ROST-P-* and ROST-PA-* and ROST-PAC-*\\
NLP & Natural Language Processing\\
BERT & Bidirectional Encoder Representations from Transformers\\
GPT & Generative Pre-trained Transformer\\ 
PPM & Prediction by Partial Matching \\
pALM & per-Author Language Model\\ 
AUC & Area Under the Curve\\
$\times_t$ & cross-topic \\
$\times_g$ & cross-genre \\
CCTA & Consumer Credit Trade Association 
\end{tabular}
}

\appendixtitles{yes} 
\appendixstart
\appendix
\section[\appendixname~\thesection]{Parameter Settings}
\label{app:param-settings}

ANN parameters are presented in Table~\ref{tab:ANN_settings}. We decided to use a fairly simple ANN architecture, using only 3 layers as we saw from the literature (e.g., \cite{ kestemont2018overview}) that simple classifiers outperformed more sophisticated approaches based on deep learning in the case of cross-domain authorship attribution. We varied the number of neurons on the hidden layer to find a suitable ANN architecture for building our classification model. 

\begin{table}[H]
    \caption{ANN parameters.}
  \label{tab:ANN_settings}  
  \begin{tabularx}{\textwidth}{l l}
\toprule
    \textbf{Parameter} & \textbf{Value}\\
    \midrule
     Activation function & SIGMOID\\
     Maximum number of training epochs & 500\\
     Number of layers & 3 (1 input, 1 hidden, and 1 output)\\
     Number of neurons on hidden layer & from 5 to 50\\ 
     Number of inputs & 56, 415, 432 (corresponding to the considered sets)\\
     Number of outputs & 10 (corresponding to authors)\\
     Error on training and validation & RMSE\\
     Error on test & percent of incorrectly classified items\\
     Desired error on validation & 0.001\\
 \bottomrule
\end{tabularx}
\end{table}


MEP parameters are detailed in Table~\ref{tab:MEP_settings}. These parameters were obtained mostly through experimentation. We thought that small errors on the training set would also lead to small errors on the test set. However, we were wrong: the main problem we encountered was overfitting and poor generalization ability of the model. Thus, other sets of parameters can also generate similar results on the test set even if the training error will be higher.

\begin{table}[H]
    \caption{MEP parameters.}
  \label{tab:MEP_settings}
    \begin{tabularx}{\textwidth}{l l}
\toprule
    \textbf{Parameter} & \textbf{Value}\\
    \midrule
     Subpopulation size & 300\\
     Number of subpopulations & 25\\
     Subpopulations architecture & ring\\
     Migration rate & 1 (per generation)\\ 
     Chromosome length&200\\
     Crossover probability&0.9\\
     Mutation probability&0.01\\
     Tournament size&2\\
     Functions probability&0.4\\
     Variables probability&0.5\\
     Constants probability&0.1\\
     Number of generations&1000\\
     Mathematical functions&+,$-$,*, /, a$<$0?b:c, a$<$b?c:d\\
     Number of constants&5\\
     Constants initial interval&randomly generated over [0, 1]\\
     Constants can evolve?&YES\\
     Constants can evolve outside the initial interval?&YES\\
     Constants delta& 1\\
\bottomrule
\end{tabularx}
\end{table}

\begin{figure}[H]
  \centering

   \begin{subfigure}[b]{0.48\textwidth}
     \centering
     \includegraphics[width=\textwidth]{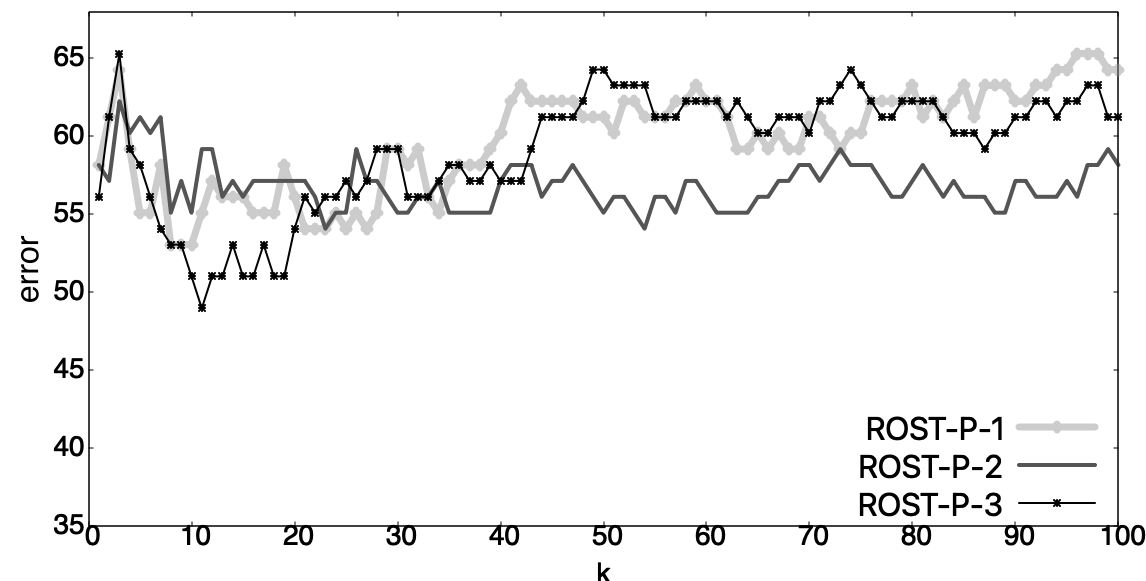}
   \end{subfigure}
   \hfill
   \begin{subfigure}[b]{0.48\textwidth}
     \centering
     \includegraphics[width=\textwidth]{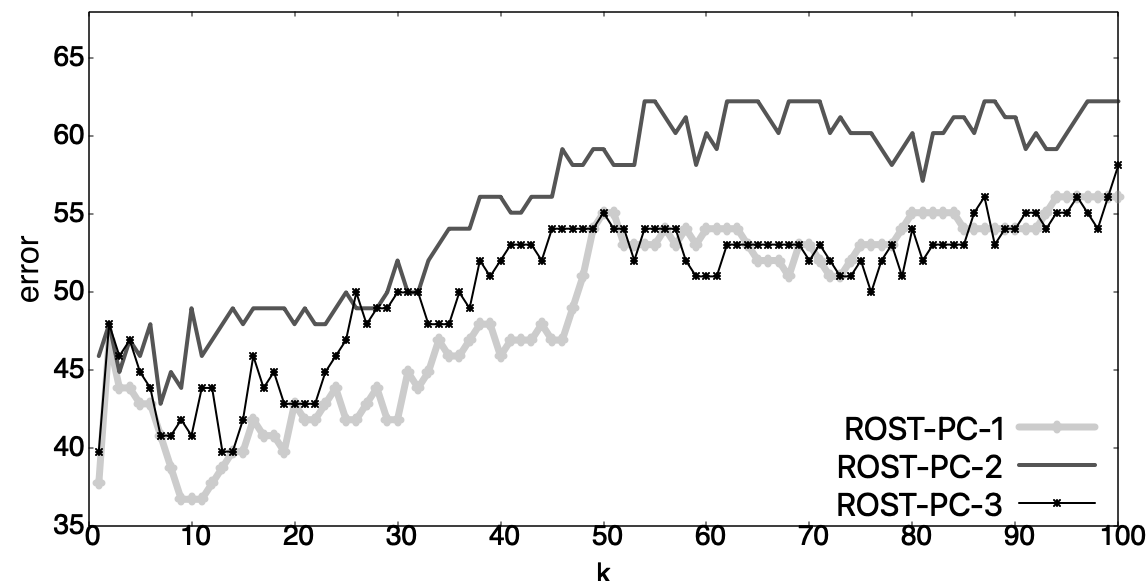}
   \end{subfigure}
      \caption{Evolution of error in k-NN for k values from 1 to 100.} 
    \label{fig:chooseK}
\end{figure}

The k-NN considers only training and test data. Thus, we have training sets of 302 items, while the test contains 98 items. During the tests, we varied the value of $k$ from 1 to 30. This is because we observed (as is depicted in Figure~\ref{fig:chooseK}) that with higher values we would not obtain better results, as the results tend to deteriorate as the value of $k$ increases. However, this depends on the number of features, as the results become bad faster for a consistent number ($>$100) of features, as for ROST-PC-*, compared to the evolution of the results for ROST-P-*, where the results do not deteriorate so fast by increasing the value of $k$ in the case of a smaller number ($<$100) of features. To calculate the distance between the test value and the ones in the training set, we used Euclidean distance. 

Support vector machines also consider only training and test data. Therefore, the training sets consist of 302 items, while the test sets contain 98 items. We experimented with the \emph{type of kernel} and \emph{nu} parameters, selecting values that varied through all possible kernel types and values from 0.001 to 1 for \emph{nu}. For the \emph{type of kernel}, the best results were obtained for linear. For \emph{nu} we had different values that gave better results depending on the dataset. However, even though we tried with values starting from 0.001, the best results were obtained for $nu\ge0.2$. We also changed the seed for the random function with no effect on the results. The SVM parameters are given in Table~\ref{tab:SVM_settings}.
\begin{table}[H]
    \caption{SVM parameters.}
  \label{tab:SVM_settings}
  \begin{tabularx}{\textwidth}{l l}
\toprule
    \textbf{Parameter} & \textbf{Value}\\
    \midrule Type of SVM & nu-SVC\\
    Type of kernel function & linear\\
    Degree in kernel function & 3 \\
    Gamma in kernel function & $1/num\_features$ \\ 
    Coef0 in kernel function & 0 \\
    nu& from 0.1 to 1 \\
    Cache memory size in MB&100 \\
    Tolerance of termination criterion (\emph{epsilon})&0.001 \\
    Shrinking heuristics, 0 or 1 & 1 \\
    Whether to train an SVC model for probability estimates, 0 or 1 & 0 \\
\bottomrule
\end{tabularx}
\end{table}

 As with k-NN and SVM, the decision trees with the C5.0 algorithm also use only training and test data. Thus, there are 302 items in the training sets and 98 items in the tests. 
All considered features/attributes, which in our case are: prepositions, adverbs, and conjunctions, are set for the Decision Trees with C5.0 
as ``\emph{continuous}'' because they are numerical float values between 0 and 1 representing the frequency of occurrence in terms of the total number of words in the file in which they occur. For authors, we set explicit-defined discrete values from 0 to 9 for the 10 authors (as specified in the first columns of Tables~\ref{tab:textStats2}
--\ref{tab:textDiv}). To improve our results, we experimented with advanced pruning options. These parameters along with others we used for decision trees with C5.0 are shown in Table~\ref{tab:DT_settings}. Apart from these parameters we also used the option $-I$ $seed$, to set the random seed, with $seed \in \{1,2,\dots,9,10,20, \dots,100\}$, without any effect on the results.
\begin{table}[H]
    \caption{Parameters for decision trees with C5.0.}
  \label{tab:DT_settings}
  \begin{tabularx}{\textwidth}{l l}
\toprule
    \textbf{Parameter} & \textbf{Value}\\
    \midrule
    No. of attributes & 57, 416, 433 (corresponding to the considered \\
    & sets plus one more attribute for the author )\\
    Global tree pruning & \emph{w} and \emph{w}/\emph{o} (option \emph{$-$g})\\
    Pruning confidence & option \emph{$-$c} $CF$, with $CF \in \{10,20,\dots,100\}$\\
    Minimum 2 branches for $\ge$ \emph{cases}& option \emph{$-$m} $cases$, with $cases \in \{1,2,\dots,30\}$\\
\bottomrule
\end{tabularx}
\end{table}

\section[\appendixname~\thesection]{Prerequisite Tests and Results}
\label{app:prereq-results}
\subsection[\appendixname~\thesubsection]{ANN}
\label{app:prereq-results-ANN}
\begin{figure}[H]
  
  \includegraphics[width=.6\textwidth]{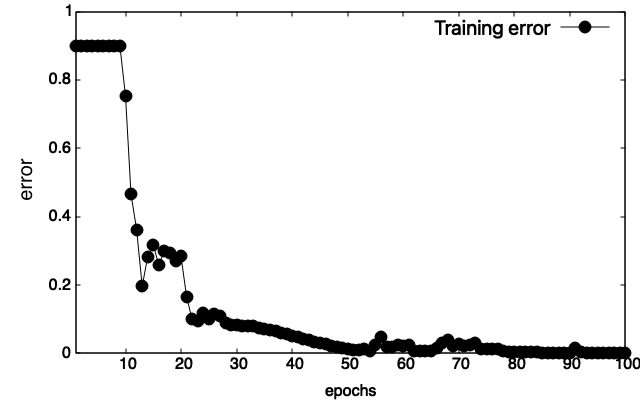}
  \caption{ANN error evolution on a training set.}
  \label{fig:ANN-trainignErrEvol}
\end{figure}

As a prerequisite, we are interested in seeing how ANN evolves while training on the data. The ANN error evolution on a training set is shown in Figure~\ref{fig:ANN-trainignErrEvol}. 

It can be seen here that within 20 epochs, the training error drops below 0.1, while within 60 epochs, it reaches 0. Thus, ANN can be used to solve this kind of problem.

Next, we want to find a good value for the number of neurons on the hidden layer. In total, 30 runs were performed with the number of neurons (on the hidden layer) varying from 5 to 50. 
 
\begin{figure}[H]
  
  \hspace{-10pt} \begin{subfigure}[b]{\textwidth}
     
     \includegraphics[width=0.32\textwidth]{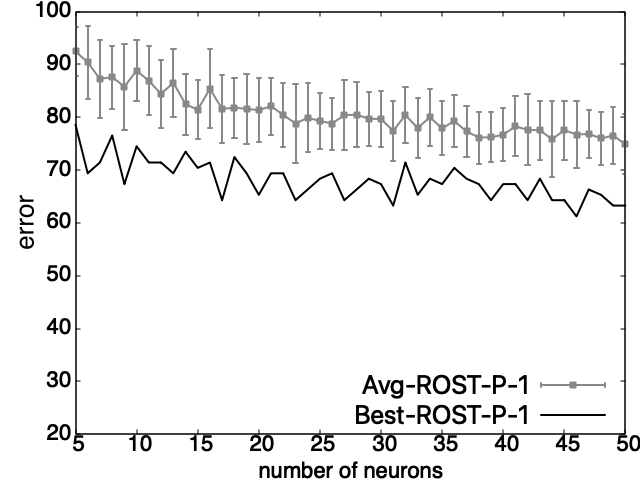}
     \includegraphics[width=0.32\textwidth]{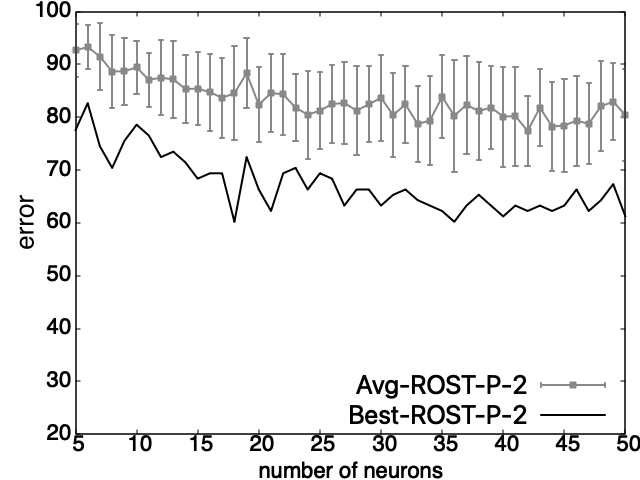}
     \includegraphics[width=0.32\textwidth]{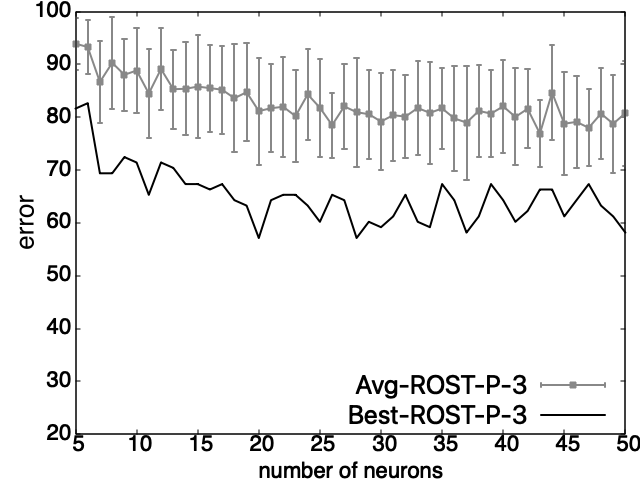}
   \end{subfigure}
   \hfill
   \hspace{-10pt}\begin{subfigure}[b]{\textwidth}
   
     \includegraphics[width=0.32\textwidth]{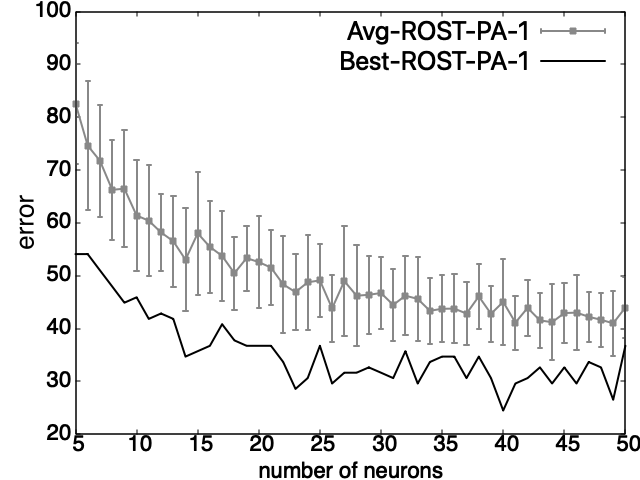}
     \includegraphics[width=0.32\textwidth]{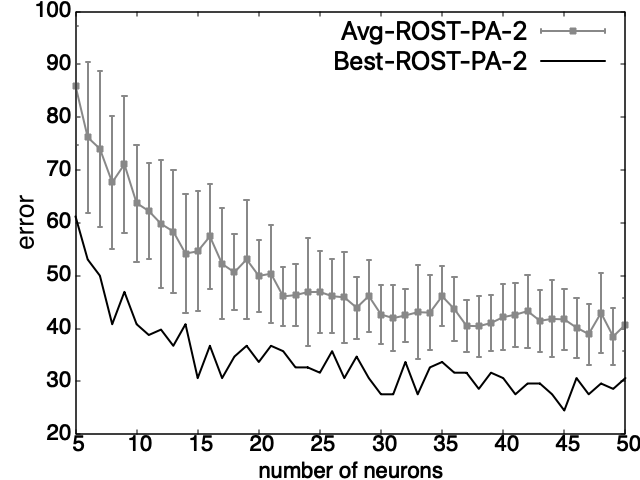}
     \includegraphics[width=0.32\textwidth]{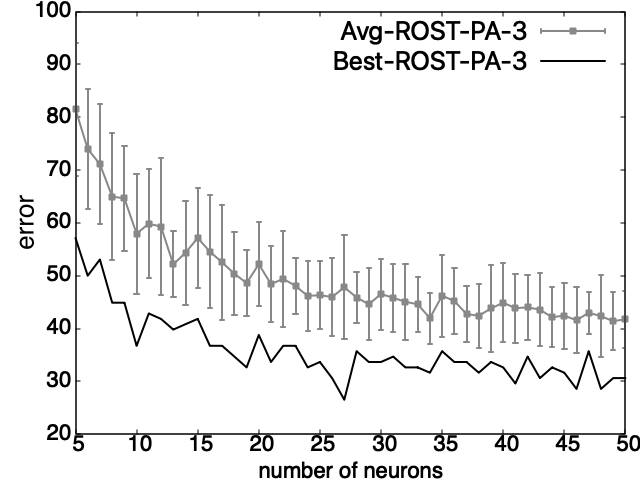}     
   \end{subfigure}
   \hfill
   \hspace{-10pt}\begin{subfigure}[b]{\textwidth}
   
     \includegraphics[width=0.32\textwidth]{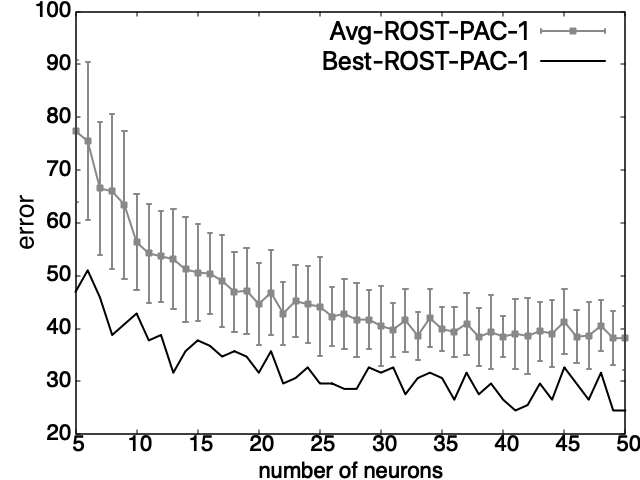}
     \includegraphics[width=0.32\textwidth]{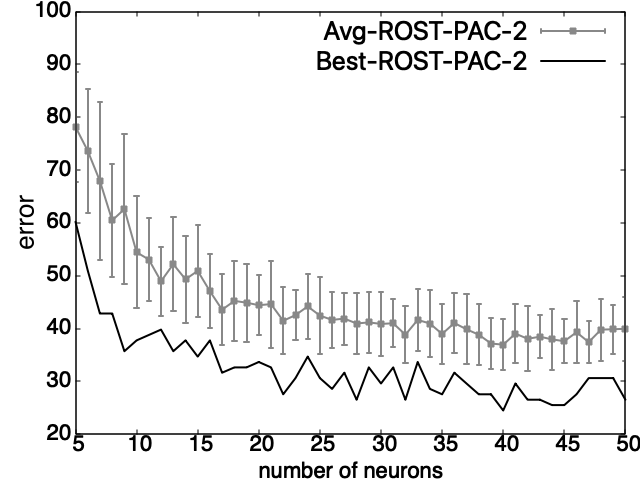}
     \includegraphics[width=0.32\textwidth]{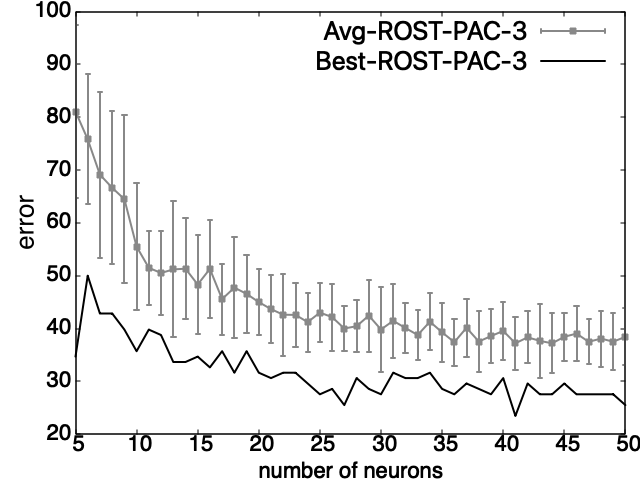}
   \end{subfigure}
    \caption{ANN results on the considered datasets. On each set, 30 runs are performed by ANNs with the hidden layer containing from 5 to 50 neurons. The percentage of incorrectly classified data is plotted. \textit{Best} stands for the best solution (out of 30 runs), \textit{Avg} stands for \textit{Average} (over 30 runs), and the \textit{Standard Deviation} is represented by error bars.} 
    \label{fig:fann-rez}
\end{figure}

The results for the 9 ROST-*-* are presented in Figure~\ref{fig:fann-rez}. 

These graphics show that, using only 56 features (i.e., ROST-P-*) tests errors were very high, while with an increased number of features: i.e., 415 (i.e., ROST-PA-*) and 432 (i.e., ROST-PAC-*), respectively, test errors are significantly reduced. Moreover, it appears that the test error values tend to stabilize between 40 and 50 neurons on the hidden layer. 

\subsection[\appendixname~\thesubsection]{MEP}
\label{app:prereq-results-MEP}
\begin{figure}[H]
 
  \hspace{-10pt}\includegraphics[width=.75\textwidth]{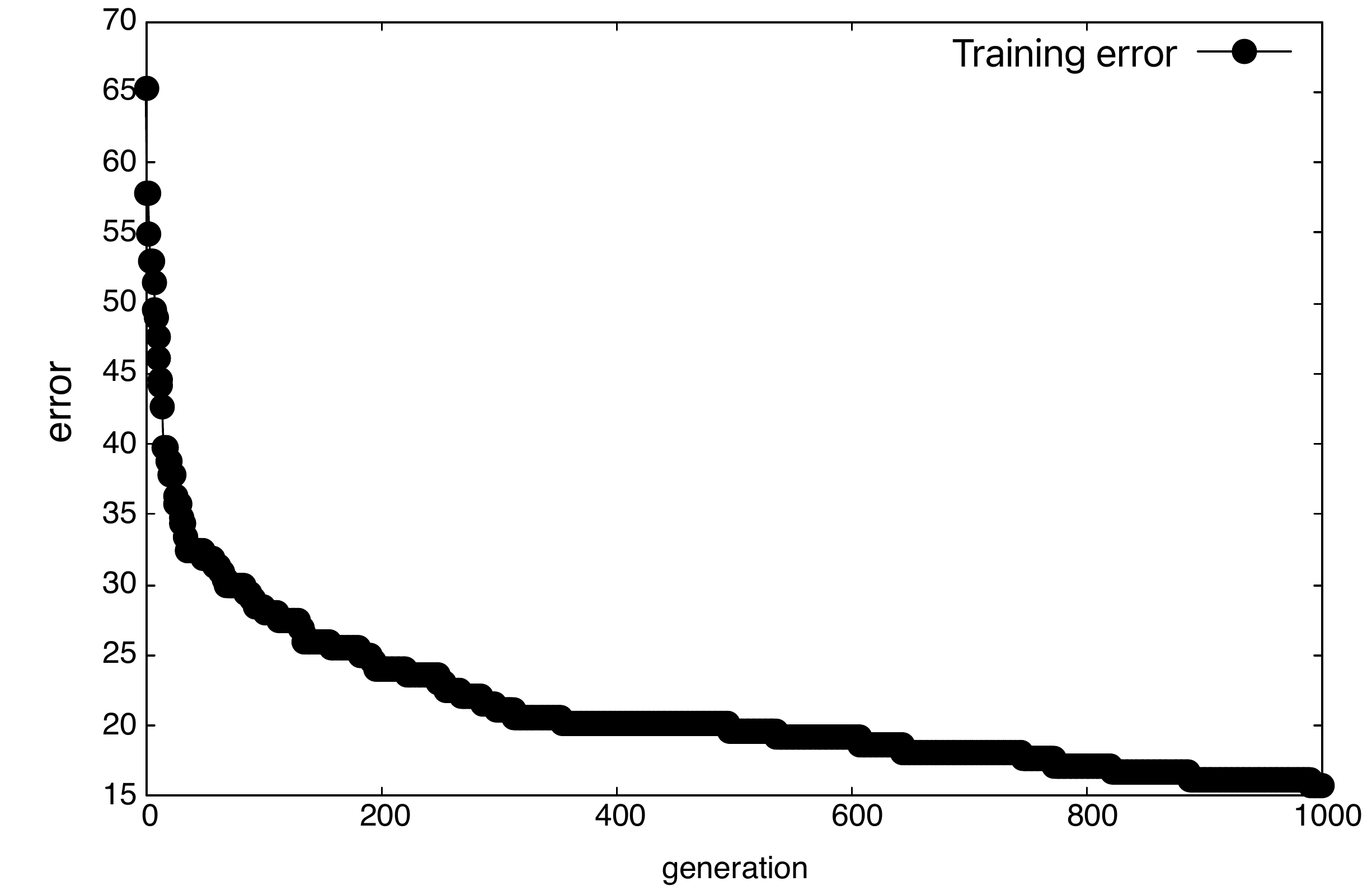}
  \caption{MEP error evolution on a training set.}
  \label{fig:MEP-trainignErrEvol}
\end{figure}

We are interested to see if MEP is able to discover a classifier and then to see how well it performs on new (test) data. The evolution of MEP error on a training set is shown in Figure~\ref{fig:MEP-trainignErrEvol}. One can see that the error rapidly drops from over 65\% to 15\%. This means that MEP can handle this type of problem.

\subsection[\appendixname~\thesubsection]{K-NN}
\label{app:prereq-results-kNN}
\begin{figure}[H]
   \begin{subfigure}[b]{0.32\textwidth}
     \centering
     \includegraphics[width=\textwidth]{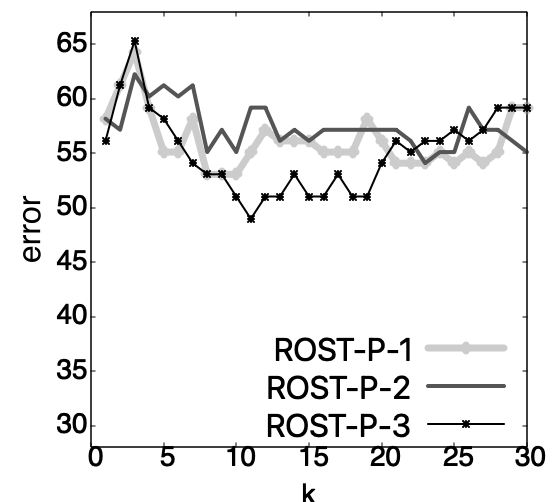}
   \end{subfigure}
   \hfill
   \begin{subfigure}[b]{0.32\textwidth}
     \centering
     \includegraphics[width=\textwidth]{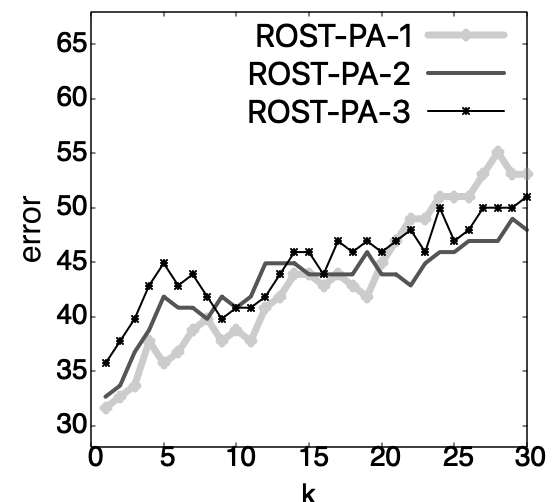}
   \end{subfigure}
   \hfill
   \begin{subfigure}[b]{0.32\textwidth}
     \centering
     \includegraphics[width=\textwidth]{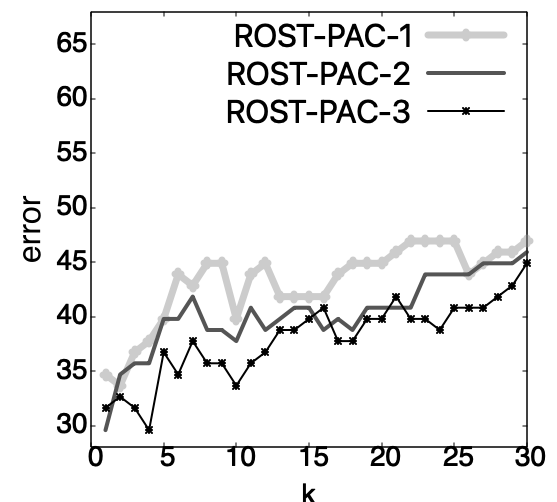}
   \end{subfigure}
    \caption{K-NN results on the considered datasets. In total, 30 runs are performed with $k$ varying with the run index. The percentage of incorrectly classified data is plotted.} 
    \label{fig:kNN-rez}
\end{figure}

With k-NN we ran tests with k varying from 1 to 30. The results for all 9 ROST-*-* are plotted in Figure~\ref{fig:kNN-rez}. It can be seen that the results for the 3 ROST-P-* 
have worst values than the values obtained for ROST-PA-* or ROST-PAC-*. 

\subsection[\appendixname~\thesubsection]{SVM}
\label{app:prereq-results-SVM}
\begin{figure}[H]
 
  \hspace{-3pt} \includegraphics[width=.32\textwidth]{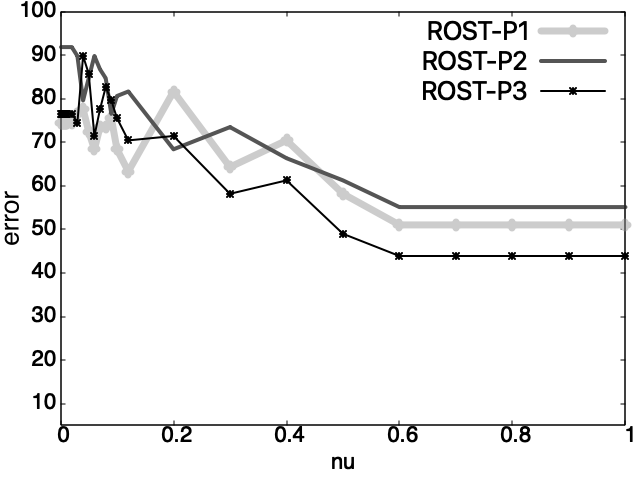}
  \includegraphics[width=.32\textwidth]{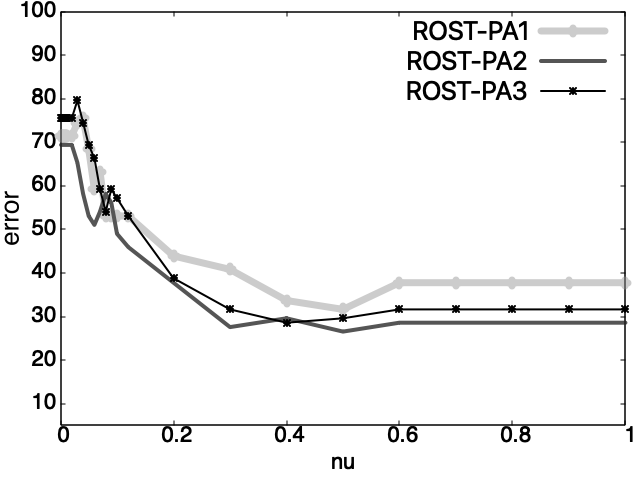}
  \includegraphics[width=.32\textwidth]{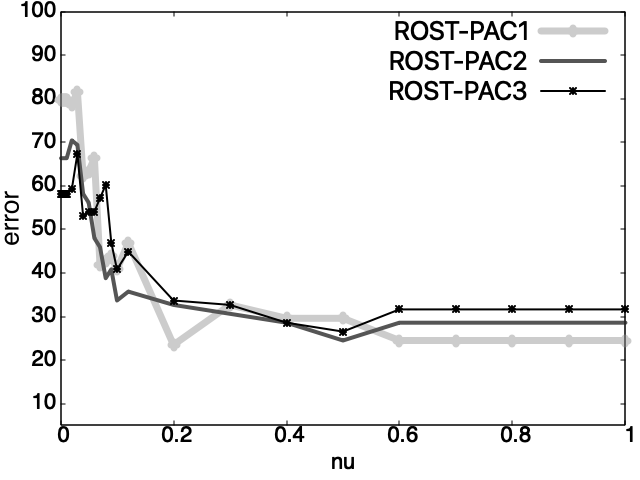}

  \caption{SVM results on the considered datasets. In total, 30 runs are performed with $nu$ varying from 0.001 to 1. The percentage of incorrectly classified data is plotted.}
    \label{fig:SVM-nu}
\end{figure}

Initially, we tried to obtain the best kernel type for our tests, and as we have already read in the literature (e.g., in~\cite{pavelec2008using}) it seems that the \emph{linear} type is the best for these types of problems (i.e., the classification for authorship attribution). We obtained significantly better results for this type as well. 
With this kernel type, we tried the find the best value for the \emph{nu} parameter. Therefore, we run tests on all our ROST-*-* with \emph{nu} values between 0.001 to 1. The results are shown in Figure~\ref{fig:SVM-nu}.

\subsection[\appendixname~\thesubsection]{DT}
\label{app:prereq-results-DT}
To optimize this method, we tried the advanced pruning options. For this we tried 3 options: 
\begin{itemize}
  \item $-g$, which disables the global tree pruning mechanism that prunes parts (of an initially large tree) that are predicted to have high error rates. 
  \item $-c$ $CF$, changes the estimation of error rates. This affects the ``severity of pruning''. $CF$ stands for \emph{confidence level} and is a percentage. We chose values from 10 to 100 for the $CF$ parameter. 
  \item $-m$ $cases$, which influences the construction of the decision tree by having at least 2 branches at each branch point for which there are more than $cases$ training items. The default value for $cases$ is 2. We have selected values from 1 to 30 for the $cases$ parameter. 
\end{itemize}

 The results obtained using decision trees with C5.0 are detailed in Table~\ref{tab:rez-DT}. 
\begin{table}[H]
\caption{Decision tree 
 results on the considered datasets. The number of incorrectly classified data is given as a percentage. Result sets are grouped into columns of \emph{Error}, \emph{Size}, and sometimes a parameter. The first set of \emph{Error}, \emph{Size} columns represent the results obtained with no options. \emph{$-$g} stands for global tree pruning is disabled, \emph{$-$c CF} stands for setting the confidence level via the \emph{CF} parameter, and \emph{$-$m cases} stands for controlling how the decision tree is built by using the \emph{cases} parameter. \emph{Error} stands for the test error rate, \textit{Size} stands for the size of the decision tree required for that specific solution, \emph{CF} stands for ``confidence level'' ($CF \in \{10, 20, \dots, 100\}$), and \emph{cases} stands for the threshold for which is decided to have two more that two branches at a specific branching point ($cases \in \{1, 2, \dots, 30\}$).\label{tab:rez-DT}}
	\begin{adjustwidth}{-\extralength}{0cm}
		\begin{tabularx}{\fulllength}{Ccccccccccc}
		\toprule
 & \multicolumn{2}{c}{}& \multicolumn{2}{c}{ \boldmath{$-g$} }& \multicolumn{3}{c}{ \textbf{\boldmath{$-c$} CF}}& \multicolumn{3}{c}{ \textbf{\boldmath{$-m$} Cases}}\\
\textbf{Dataset} & \textbf{Error}& \textbf{Size}& \textbf{Error}& \textbf{Size}& \textbf{Error}& \textbf{Size}&\textbf{CF}& \textbf{Error}& \textbf{Size}&\textbf{Cases}\\ \midrule
\mbox{ROST-P-1} &58.2\%& 60 & 58.2\%& 60& 58.2\%& 60 & $\ge$10&51.0\%& 18 & 8\\
\mbox{ROST-P-2} &53.1\%& 57 & \boxed{54.1\%}& \boxed{61}& 53.1\%& 57 & $\ge$10&51.0\%& 46 & 3\\
\mbox{ROST-P-3} &69.4\%& 64 & 69.4\%& 64& 69.4\%& \boxed{56} & \boxed{=10}&57.1\%& 99 &1\\\midrule
\mbox{ROST-PA-1} &35.7\%& 39 & 35.7\%& \boxed{42}& 35.7\%& 39 & $\ge10$&31.6\%& 13 &12\\
\mbox{ROST-PA-2} &28.6\%& 38 & \boxed{27.6\%}& \boxed{42}& \boxed{27.6\%}& \boxed{43} & \boxed{>20}&26.5\%& 57 &1\\
\mbox{ROST-PA-3} &30.6\% & 38 & 30.6\%& \boxed{40}& 30.6\%& 38 & $\ge$10&29.6\%& 31 &3\\\midrule
\mbox{ROST-PAC-1} &28.6\%& 39 & 28.6\%& \boxed{41}& 28.6\%& 39 & $\ge$10&28.6\%& 39 &2\\
\mbox{ROST-PAC-2}&25.5\% & 37 & 25.5\%& \boxed{39}& 25.5\%& 37 & $\ge$10& 24.5\%& 12 &14\\
\mbox{ROST-PAC-3} &32.7\% & 38 & \boxed{33.7\%}& \boxed{41} & 32.7\%& 38 & $\ge$10&26.5\%& 13 &14\\
\bottomrule
		\end{tabularx}
	\end{adjustwidth}
\end{table}
Using the $-g$ option, it can be seen that most of the trees have become larger (as expected since global tree pruning was disabled by this option). Changes in the results are marked in the table with the values in the boxes. However, most results remained the same, two worsened (i.e., for ROST-P-2 from 53.1\% to 54.1\% and for ROST-PAC-3 from 32.7\% to 33.7\%), while only one result improved (i.e., for ROST-PA-2 from 28.6\% to 27.6\%).

When using the $-c$ CF option, almost all results were similar to those obtained without using any option. The exceptions (marked with boxes), i.e., for ROST-PA-2 better results (i.e., 27.7\% vs. 28.6\% and the same as using the $-g$ option) were obtained by using a larger tree (i.e., 43 compared to 38; in this case larger than using the $-g$ option, which had a tree size of 42). In the case of ROST-P-3, only the tree size was optimized from 64 to 56 for $CF=10$. For $CF>20$, both the error and the tree size remained the same as without using any option, while for $CF=20$, the tree size was slightly reduced (i.e., 63 from 64), but the error was higher (i.e., 70.4\% from 69.4\%).

Using the $-m$ cases option, we obtained improvements, as shown in Table~\ref{tab:rez-DT}. All error rates were improved, while the improvement in the tree size, although in some cases was significant (i.e., from 60 to 18, from 39 to 13, from 37 to 12, or from 38 to 14), in other cases the tree size increased or remained large (i.e., for ROST-P-3 from 64 to 99, for ROST-PA-2 from 38 to 58, and for ROST-PAC-1 it remained the same as when no option was used). For these three ROST-*-* mentioned above for which the tree size increased or remained large, the value of the \emph{cases} parameter was very low (i.e., 1 or 2). For ROST-P-2 and ROST-PA-3, there is $cases=3$ and the tree size did not change that much (i.e., from 38 to 31) and remained the same, respectively. For ROST-P-1, ROST-PA-1, ROST-PAC-2, and ROST-PA-3, $cases\ge 8$, and the three decision trees have greatly reduced in size to values~$\le$18.

\begin{figure}[H]
 
  \includegraphics[width=.32\textwidth]{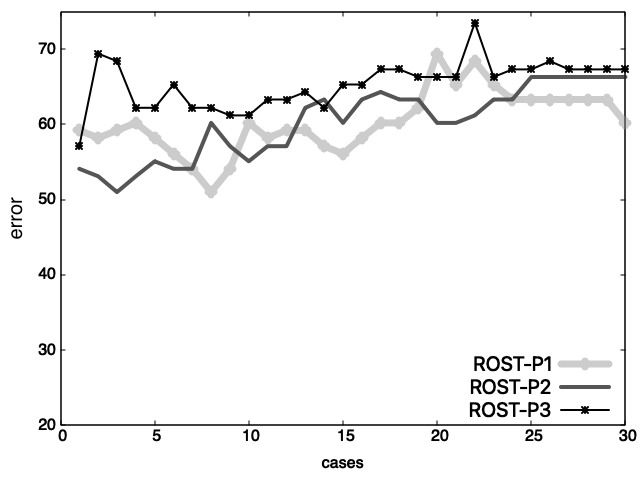}
  \includegraphics[width=.32\textwidth]{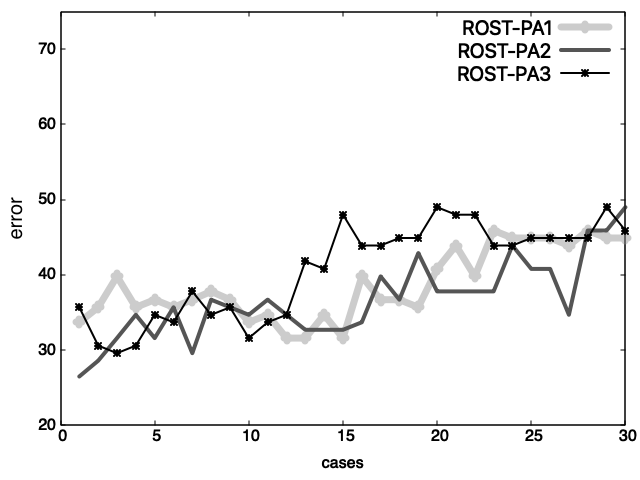}
  \includegraphics[width=.32\textwidth]{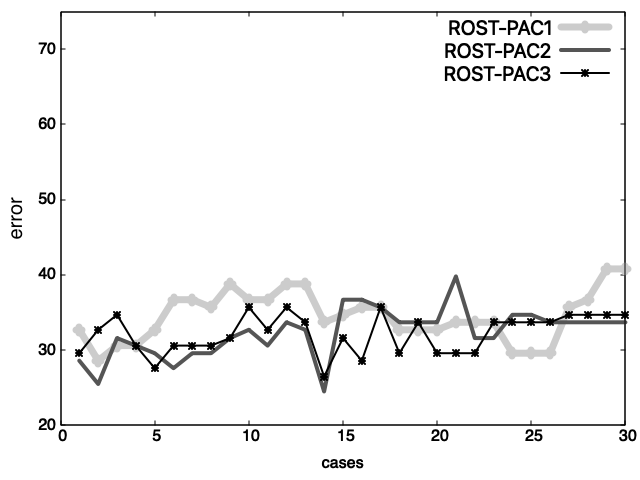}

  \caption{DT results on the considered datasets. In total, 30 runs are performed with the $cases$ parameter (introduced by the $-m$ option) varying from 1 to 30. The percentage of incorrectly classified data is plotted.} 
    \label{fig:DT-m}
\end{figure}

To show the evolution of the error rates for the three datasets considered, we plotted the results of the decision trees obtained using C5.0 with the $-m$ option, with $cases$ varying from 1 to 30. The results are shown in Figure~\ref{fig:DT-m}.

\begin{adjustwidth}{-\extralength}{0cm}

\reftitle{References}

\end{adjustwidth}
\end{document}